\documentclass[11pt]{article}
\usepackage{amsfonts}
\usepackage{amsmath}

\usepackage[final]{acl}

\usepackage{times}
\usepackage{latexsym}
\usepackage{array}
\usepackage{tabularx}
\usepackage{colortbl}  
\usepackage{xcolor}
\usepackage{colortbl}
\usepackage{float}
\usepackage{soul}
\setulcolor{red}
\usepackage[dvipsnames]{xcolor}

\usepackage[T1]{fontenc}

\usepackage[utf8]{inputenc}

\usepackage{microtype}

\usepackage{inconsolata}

\usepackage{graphicx}
\usepackage[utf8]{inputenc}
\usepackage{hyperref}
\usepackage{url}
\usepackage{booktabs,multirow}
\usepackage{amssymb}
\usepackage{pifont}
\usepackage{adjustbox}
\usepackage{enumitem}
\usepackage{wrapfig}
\usepackage{xcolor}

\usepackage{pgfplots}
\usetikzlibrary{pgfplots.groupplots,intersections,pgfplots.fillbetween}
\pgfplotsset{compat=1.3}
\usepackage{scalefnt} 

\title{\textit{Everyone is unique}: Towards Behaviorally Heterogeneous Negotiation Dialogue Systems for Debt Collection}

\author{
\textbf{Yuhang Yang\textsuperscript{1,2}}, 
\textbf{Kai Tang\textsuperscript{1,2}},
\textbf{Chao Ye\textsuperscript{1}}, 
\textbf{Haobo Wang\textsuperscript{1}\thanks{Corresponding authors.}},
\\
\textbf{Qiqi Luo\textsuperscript{2}\footnotemark[1],} 
\textbf{Jinguang Zheng\textsuperscript{2},}
\textbf{Zhixin Zhang\textsuperscript{2}}\\
\textsuperscript{1}State Key Laboratory of Blockchain and Data Security, Zhejiang University \\
\textsuperscript{2}Ant Group \\
\{yangyuhang, wanghaobo\}@zju.edu.cn \\
\{luoqiqi.lqq, zhengjinguang.zhen\}@antgroup.com \\
}

\definecolor{darkgreen}{rgb}{0.0,0.5,0.0}

\begin{document}
\maketitle
\begin{abstract}
Debt collection is a critical negotiation task in the financial industry, with strong practical relevance and exceptional academic value as a behaviorally rich, high-stakes testbed for human-centered dialogue systems. While large language models (LLMs) have shown promise in dialogue and negotiation, effectively evaluating their performance in this complex scenarios remains a major challenge: existing benchmarks uniformly assume users to be static, rational agents with fixed preferences, failing to capture the rich behavioral heterogeneity inherent in real-world debt collection. To bridge this gap, we propose \textbf{DebtBench}, the first public persona-enriched debt collection benchmark, that highlights behavioral heterogeneity in negotiation. Moreover, we develop \textbf{DebtGPT}, a debt collection agent trained to jointly optimize financial recovery and interaction experience. Our experimental results, using \textbf{16} state-of-the-art LLMs, find that most existing models struggle in this complex but realistic scenarios, whereas DebtGPT outperforms all open-source baselines and achieves performance on par with GPT-4o. The code and data are available at \url{https://github.com/YYuHhhh/DebtNegotiation}.
\end{abstract}

\begin{figure}[t]
\centering
\includegraphics[width=\linewidth]{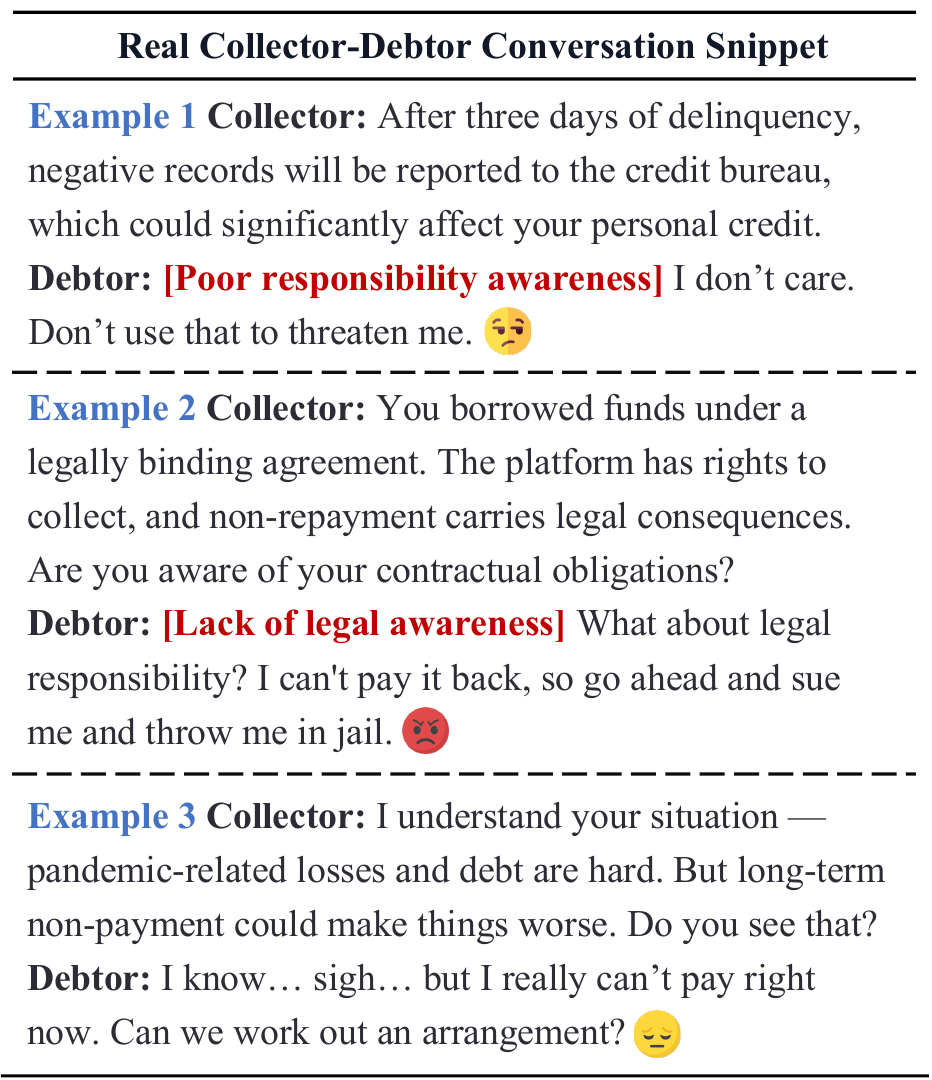}
\caption{Real-world collector-debtor dialogues illustrating three key behavioral characteristics: \textbf{(1) Rich emotional expression}, \textbf{(2) Cognitive limitations}, and \textbf{(3) Diverse linguistic styles.}}
\label{fig:debtor_conversation}
\end{figure}

\section{Introduction}

Debt collection is a critical yet highly labor-intensive task in the financial industry, where institutions must recover non-performing loans while preserving debtor engagement and compliance. Each year, millions of individuals fall into prolonged delinquency due to personal financial hardship, necessitating negotiation-based resolution to minimize creditor losses and avoid legal escalation \citep{ozili2019non, firanda2021debt}. Traditionally, the process relies on large teams of professional collectors to manually contact debtors, negotiate repayment terms, and follow up on overdue accounts \citep{yang2020personalizing, abe2010optimizing}. This labor-intensive approach incurs substantial operational costs and is prone to human error—ultimately resulting in suboptimal recovery rates and poor customer experiences. These limitations underscore a pressing need for automated negotiation agents capable of scaling personalized, effective, and empathetic debt resolution.

%

Recent advances in large language models (LLMs) and associated benchmarks \citep{Bargainbench, FishBargain} have advanced negotiation agents, but they focus on simple, well-structured scenarios like item bargaining and e-commerce haggling. While \citet{wang2025debt} extend negotiation to complex domains such as debt collection, they oversimplify the scenario by modeling only objective financial attributes but overlooking subjective behavioral factors. Critically, all these benchmarks uniformly assume users are static, rational agents with fixed preferences, failing to capture the behavioral heterogeneity observed in practice. As illustrated in Figure~\ref{fig:debtor_conversation}, real-world debt negotiation exhibits three salient behavioral heterogeneities that challenge these idealized models:
\textbf{(1) Rich emotional expression}—users display intense, dynamic emotions such as anxiety, defensiveness, and sadness;
\textbf{(2) Cognitive limitations}—individuals often hold legal misconceptions, lack financial literacy, or reason inconsistently; and
\textbf{(3) Diverse linguistic styles}—ranging from evasive, fragmented utterances to bluntly confrontational statements. These complexities—stemming from the fact that \textit{everyone is unique}—render existing benchmarks inadequate for debt collection.

Along with its high practical relevance, debt collection offers exceptional academic value as a behaviorally rich, high-stakes testbed for human-centered negotiation agents. However, collecting authentic dialogues at scale is extremely challenging due to stringent privacy regulations governing sensitive personal and financial data, limiting prior work \citep{wang2025debt} to superficial explorations. Fortunately, through collaboration with a leading financial technology company, we gained access to a large corpus of real collector–debtor conversations. While confidentiality agreements prohibit public release of the raw data, they allow us to distill authentic behavioral patterns into a privacy-preserving synthetic benchmark. To this end, we propose \textbf{DebtBench}, the first public persona-enriched debt collection benchmark, that highlights behavioral heterogeneity in negotiation. This procedure is driven by a three-stage data synthesis pipeline, where we half-automatically extract chatting principles in real-world samples and then, prompt a leading LLM to iteratively generate high-quality, multi-dimensional persona profiles.

To advance the development of far-sighted, user-adaptive negotiation agents in debt collection, we develop \textbf{DebtGPT}, an agent trained via Coarse-to-Fine Preference Optimization (CFPO)—a framework that enables models to learn far-sighted negotiation policies that jointly optimize long-term financial recovery and user experience. We conduct a comprehensive evaluation of advanced open-source and closed-source LLMs on our DebtBench benchmark. The results reveal that: (1) Most existing models struggle in persona-enriched debt collection scenarios, achieving success rates below 75\%. (2) Contrary to expectations, reasoning-specialized models underperform their general-purpose counterparts, highlighting a misalignment between formal reasoning capabilities and the behaviorally grounded demands of real-world negotiation. (3) Notably, our 8-billion-parameter DebtGPT outperforms all open-source baselines and achieves performance comparable to GPT-4o.

\section{Related Work}

\paragraph{Negotiation Benchmarks.}
Negotiation is an active area in NLP, with increasing focus on dialogue systems. Recent benchmarks have introduced various negotiation, which can be broadly categorized into cooperative and competitive paradigms. Cooperative negotiation focuses on multi-issue trade-offs to achieve mutual gains, as seen in Persuasion \citep{wang2019persuasion, jin2024persuading} and JobInterview \citep{jobinterview}. In contrast, competitive negotiation models zero-sum haggling over fixed resources, exemplified by Bargain \citep{CraigslistBargain, Bargainbench, FishBargain} and Assignment \citep{lewis2017deal, chawla2021casino}, where one party’s gain directly reduces the other’s. Despite significant progress, existing negotiation dialogue systems mostly address simple, well-structured scenarios. Although \citep{wang2025debt} extend negotiation to complex, high-stakes domains like debt collection, they model only objective financial attributes and neglect subjective behavioral factors. Crucially, all current benchmarks assume users are static, rational agents with fixed preferences, thus failing to capture real-world behavioral heterogeneity.
\paragraph{Large Language Model in Negotiation.}
Recent research has sought to improve the strategic capabilities of large language models (LLMs) in negotiation dialogue settings, which can be categorized into two main paradigms: 
1) using prompt engineering to elicit internal reasoning. \citet{askexp} and \citet{DengLC0LC23} prompt LLMs to plan next-turn actions through self-reflection. \citet{self-play} employ self-play between LLMs to iteratively refine negotiation strategies via AI-generated feedback. 
2) introducing external planners. \citet{ppdpp} introduce a plug-and-play policy planner to generate strategic guidance for dialogue agents. \citet{DPDP} and \citet{MCTS} employ Monte Carlo Tree Search to enhance long-term planning.
However, these methods focus almost exclusively on task-level success, often overlooking the user’s interaction experience. In high-stakes contexts like debt collection, poor communication can erode trust, trigger disengagement, and ultimately harm long-term recovery \citep{crsoftware2024poorcomm}. Our work addresses this gap by jointly optimizing strategic effectiveness and human-centered interaction through behaviorally grounded personas and a far-sighted training framework (CFPO).

\begin{figure*}[t]
\centering
\includegraphics[width=\linewidth, height=10.5cm]{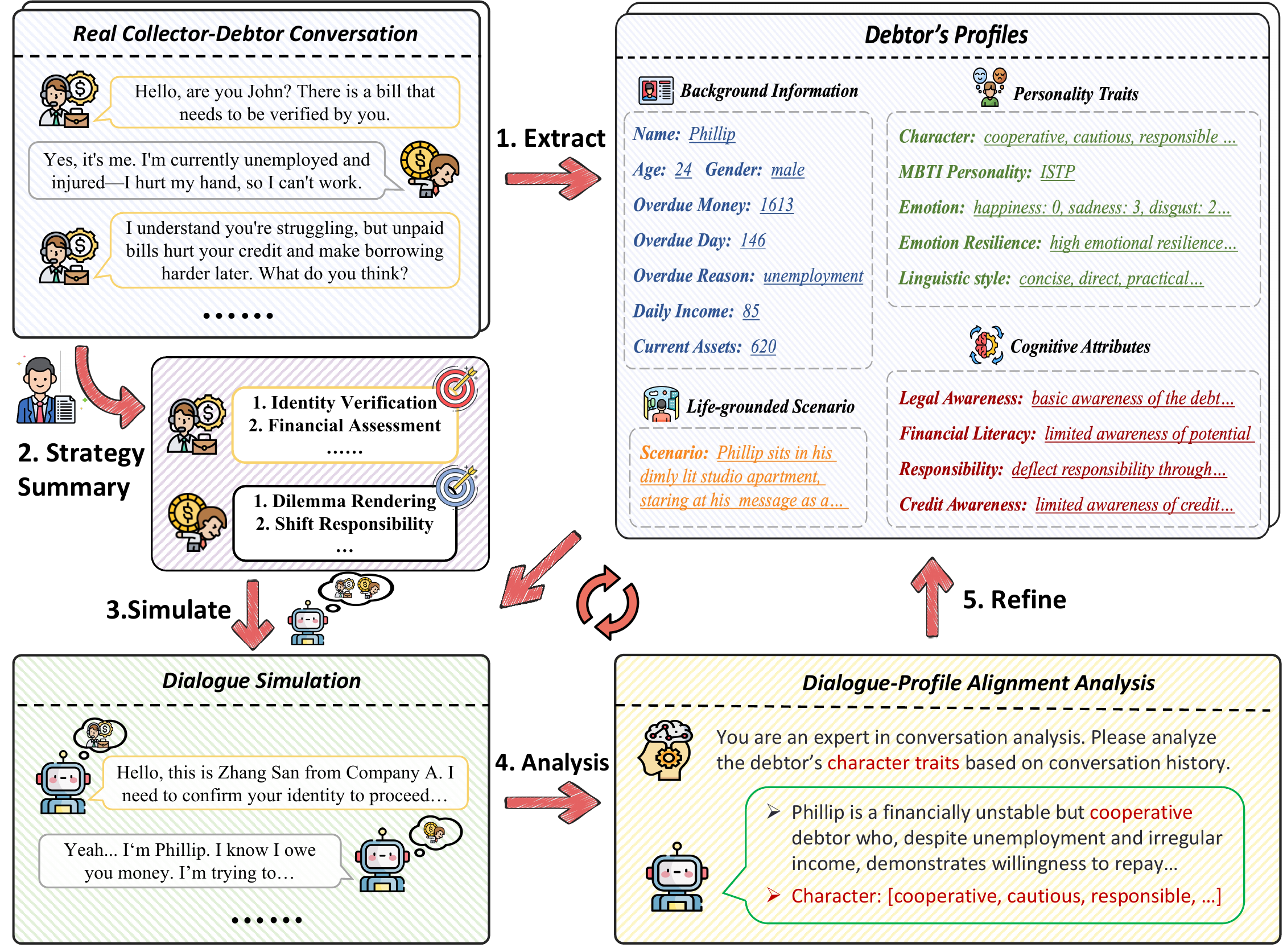}
\caption{The DebtBench Persona Synthesis Pipeline: (1) Extract multi-dimensional attributes from real dialogues; (2) Perform expert-guided strategy summarization via conversation analysis; (3-5) Iteratively refine persona profile via dialogue simulation and LLM-based alignment.}
\label{fig:dataset_construction}
\end{figure*}

\section{DebtBench}
\subsection{Data Construction}

\begin{figure}[t]
\centering
\includegraphics[width=\linewidth]{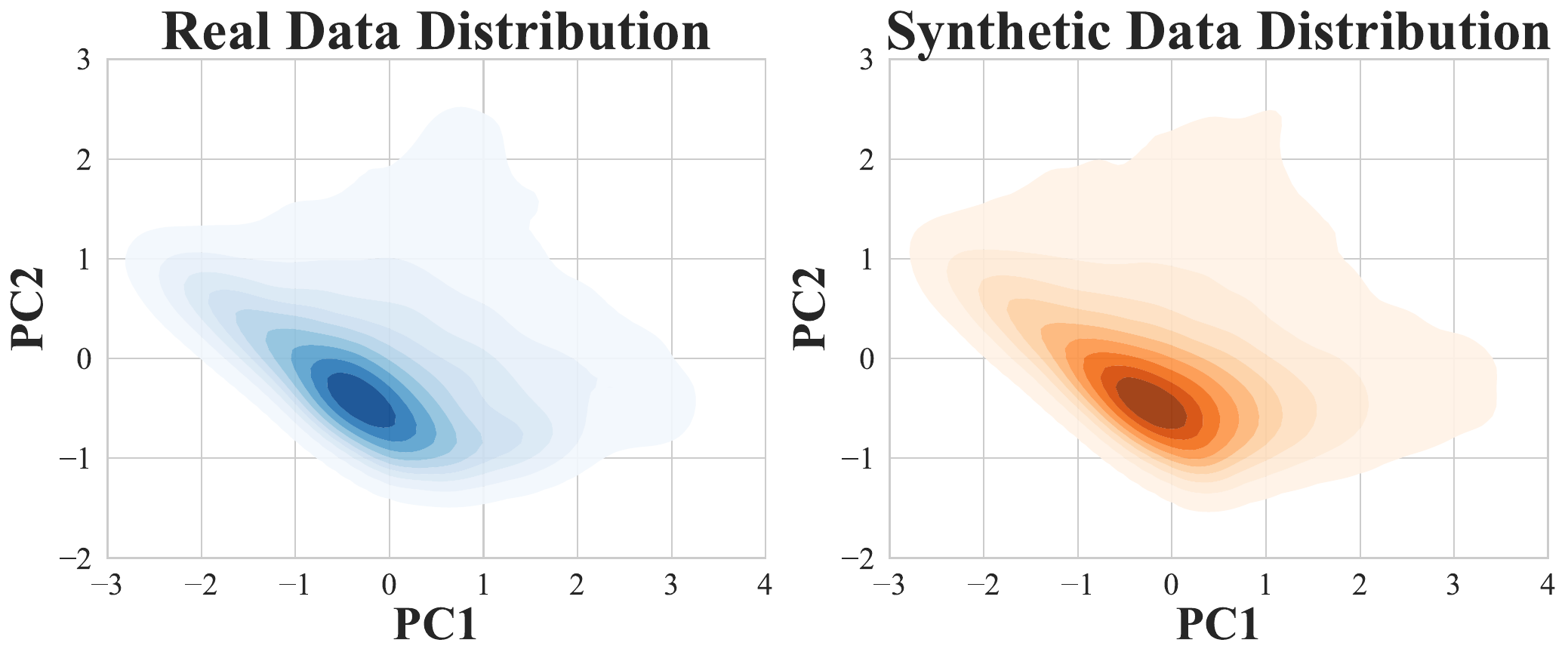}
\caption{Comparison of real and synthetic user data distributions in PCA space. Left: Real data (blue); Right: Synthetic data (orange).}
\label{fig:distribution_compariso}
\end{figure}

To advance realistic debt collection dialogue systems, we need fine-grained personas and high-quality data capturing authentic human behavior—yet such data is scarce due to privacy constraints. While we accessed real conversations from a leading fintech firm, strict confidentiality prevents public release. As shown in Figure~\ref{fig:debtor_conversation}, these dialogues reveal three core behavioral traits: \textit{1) Rich emotional expression}, \textit{2) Cognitive limitations} and \textit{3) Diverse linguistic styles}—all of which challenge LLMs’ tendency to generate overly rational, neutral responses. To bridge this gap, we propose a three-stage synthesis pipeline: \textit{Persona Extraction}, \textit{Strategy Enrichment} and \textit{Behavior Refinement}.


\paragraph{Persona Extraction.} we first derive debtor personas through systematic analysis of 1,000 real-world collector-debtor conversations provided by a leading fintech company, modeling each debtor as a unique individual along four key dimensions:
\begin{itemize}[leftmargin=2.5mm,parsep=2pt]
    \item \raisebox{-.15\height}{\includegraphics[height=1.1em]{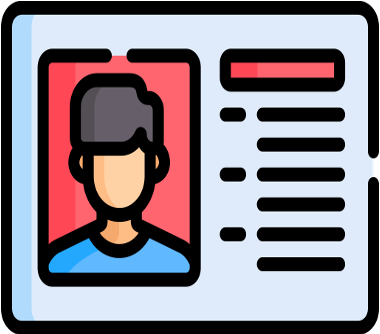}} \textbf{Background Information}: Core demographic, debt-related, and financial details that preserve the distribution of real-world debtor profiles and ensure high scenario fidelity.
    \item \raisebox{-.15\height}{\includegraphics[height=1.1em]{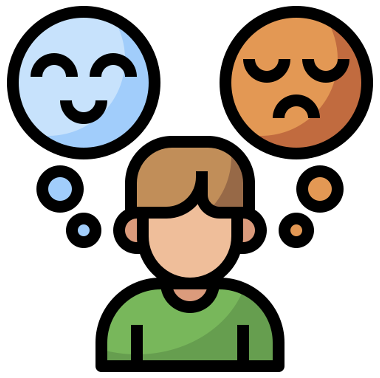}} \textbf{Personality Traits}: Affective and behavioral characteristics, including character, emotion, emotional resilience, linguistic style, and MBTI personality type—to drive \textit{rich emotional expression} and \textit{diverse linguistic styles}.
    \item \raisebox{-.15\height}{\includegraphics[height=1.1em]{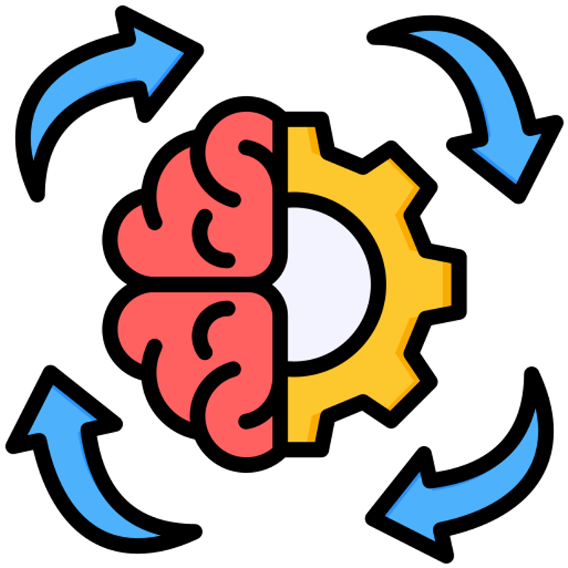}} \textbf{Cognitive Attributes}: The user’s understanding of legal obligations, financial knowledge, credit consequences, and responsibility—to reflect \textit{cognitive limitations} in real interactions.
    \item \raisebox{-.15\height}{\includegraphics[height=1.1em]{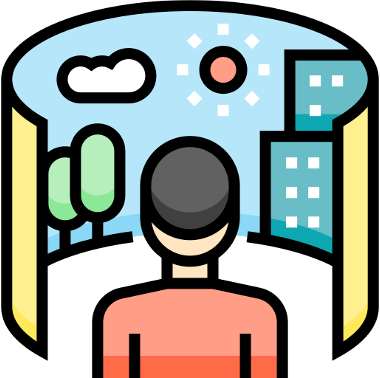}} \textbf{Life-grounded Scenario}: A coherent narrative context (e.g., job loss, medical debt) that grounds the persona in a plausible life story—to unify attributes into a consistent individual.
\end{itemize}

Each persona $\mathcal{P}_d$ in DebtBench is formally defined as a structured tuple:
$$
\mathcal{P}_d =\left( B, M, C, S \right) = \text{Extract}_\text{LLM}(d),
$$
where $d \in D_{real}$ is a real-world conversation and $\text{Extract}_\text{LLM}(\cdot)$ is a prompt-based LLM function that distills multi-dimensional attributes from dialogue data. $B , M , C , S$ denote Background, Personality, Cognition, and Scenario, respectively. We compare the distribution of real and synthetic user background information in Figure \ref{fig:distribution_compariso}, showing strong alignment in PCA space.


\paragraph{Strategy Enrichment.}
To ensure generated utterances are both realistic and strategically interpretable, we define explicit strategy sets for the collector and debtor, guided by three principles:
(1) \textit{Behavioral Grounding}: strategies reflect authentic tactics from real interactions;
(2) \textit{Semantic Distinctness}: each strategy has a clearly differentiated intent;
(3) \textit{Actionability}: strategies can be concretely realized in natural language.
We first prompt an LLM to analyze real collector–debtor dialogues and extract strategy–utterance pairs. Semantic embeddings of these pairs are clustered using HDBSCAN \citep{mcinnes2017hdbscan} to identify coherent behavioral patterns. From each cluster, 10 representative utterances are reviewed by domain experts, who iteratively refine and consolidate them into principle-aligned categories. The final set includes 9 collector strategies (e.g., \textit{financial assessment}, \textit{emotional appeasement}) and 8 debtor strategies (e.g., \textit{emotional confrontation}, \textit{complaint}). Full definitions are provided in Tables~\ref{tab:collector_strat} and~\ref{tab:debtor_strat}. 
 
\paragraph{Behavior Refinement.}
To ensure that generated dialogues faithfully reflect the behavioral patterns defined in the extracted personas, we employ an iterative refinement mechanism to perform fine-grained alignment between each utterance and the persona’s attributes. Specifically, we employ an LLM as an automated evaluator to assess alignment along three key dimensions: \textit{emotional consistency} (e.g., verifying that a highly defensive debtor expresses anxiety appropriately), \textit{cognitive plausibility} (e.g., checking that a debtor with low financial literacy avoids technical jargon), and \textit{linguistic style coherence} (e.g., ensuring the speaking style matches ``evasive'' or ``confrontational''). Responses that fail the alignment check are iteratively revised through targeted prompting, where the LLM is conditioned on the full persona to generate more consistent alternatives. This feedback loop is used to update the persona itself, closing the alignment cycle. Formally, given an initial persona $\mathcal{P}_d^{(0)}$, the refinement proceeds as:
\begin{equation}
\mathcal{P}_d^{(\ell+1)} = \text{Refine}_\text{LLM}\left( \mathcal{P}_d^{(\ell)},\ \tau^{(\ell)} \right),
\end{equation}
where $\tau$ is a dialogue trajectory simulated using persona $\mathcal{P}_d^{(\ell)}$, and $\text{Refine}_{\text{LLM}}$ adjusts persona profile to better match the behaviors exhibited in $\tau^{(\ell)}$. The process terminates when behavioral consistency is achieved or after $\ell_{\max}$ iterations.


\begin{table}[ht]
    \centering
    \begin{tabular}{llcc}
        \toprule
        \textbf{Dimension} & \textbf{Metric} & \textbf{Score} & \textbf{$\kappa$} \\
        \midrule
        \multirow{4}{*}{\textbf{Consistency}} 
        & Emotion & 4.18 & 0.54 \\
        & Cognition & 3.91 & 0.42 \\
        & Style & 4.05 & 0.43 \\
        & Scenario & 3.89 & 0.39 \\
        \midrule
        \multirow{4}{*}{\textbf{Realism}} 
        & Naturalness & 4.09 & 0.54 \\
        & Alignment & 3.96 & 0.39 \\
        & Plausibility & 4.04 & 0.52 \\
        & Fluency & 4.09 & 0.51 \\
        \bottomrule
    \end{tabular}
    \caption{Human evaluation results of DebtBench-generated dialogue quality. The $\kappa$ value \citep{fleiss1971measuring} values fall within 0.2--0.6, indicating fair to moderate inter-annotator agreement \citep{mchugh2012interrater}.}
 \label{tab:dialogue_quality}

\end{table}

\subsection{Analysis Realism of DebtBench}

To further validate the consistency and realism of DebtBench-generated dialogues, we conducted a human evaluation with 20 annotators possessing financial industry experience, who rated 200 randomly selected test-set dialogues on two core dimensions using a 5-point Likert scale: \textbf{(1) Persona Consistency}: whether the synthetic dialogue reflects the assigned user profile. \textbf{(2) Dialogue Realism}: how closely the dialogue resembles authentic debt-collection conversations. As shown in Table~\ref{tab:dialogue_quality}, the generated dialogues received consistently high scores across all sub-dimensions, indicating strong alignment with their personas and closely mirror real-world interactions. For further details on the evaluation protocol, see Appendix \ref{app:quality_human}.

\subsection{Statistics of DebtBench}

\begin{figure}[t]
\centering
\includegraphics[width=\linewidth]{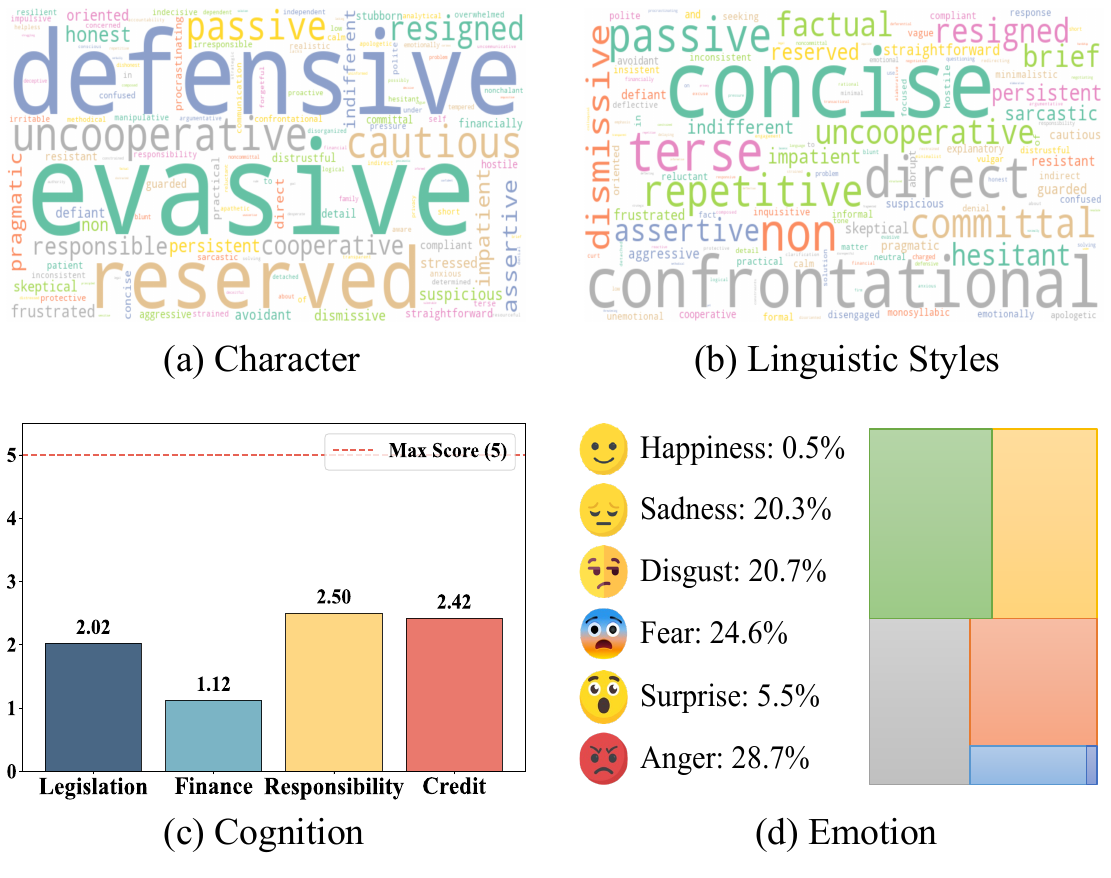}
\caption{Behavioral and Cognitive Profiles of Debtors in DebtBench. (a) and (b) are the word cloud of character traits and linguistic styles in profiles. (c) and (d) are the distribution of cognition level and emotion among profiles in the DebtBench.}
\label{fig:profile_statistics}
\end{figure}

%
Finally, we construct a high-quality dataset comprising \textbf{11,000} meticulously crafted debtor personas, split into a training set of 10,000 and a test set of 1,000 (Evaluations reported in this paper are conducted on the test set). As illustrated in Figure~\ref{fig:profile_statistics}, these personas exhibit diverse and realistic behavioral attributes across emotion, cognition, and linguistic style, enabling LLMs to generate interactions that are not only strategically grounded but also behaviorally authentic. Detailed definitions and construction methodologies for our DebtBench are provided in Appendix~\ref{appendix:debttalk_detail}.

\begin{figure}[t]
\centering
\includegraphics[width=\linewidth]{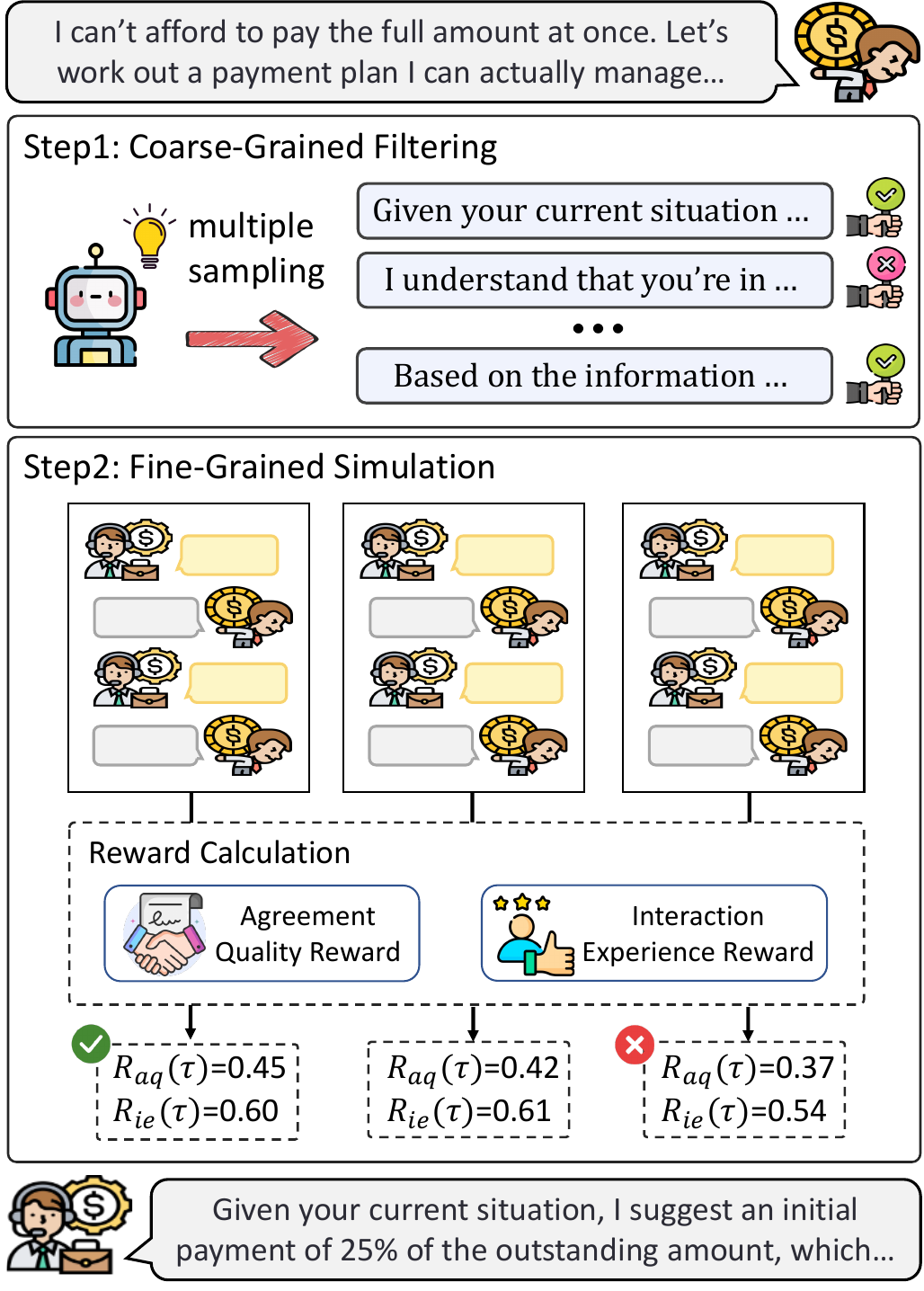}
\caption{The DebtGPT Training Framework via Coarse-to-Fine Preference Optimization (CFPO).}
\label{fig:cfpo}
\end{figure}

\section{DebtGPT}
\label{sec:method}

Building upon the behaviorally grounded DebtBench benchmark, we develop DebtGPT, a negotiation agent trained via \textit{Coarse-to-Fine Preference Optimization (CFPO)}, which combines coarse filtering with selective simulation to learn far-sighted, strategically effective, and user-adaptive policies.

\subsection{Coarse-Grained Filtering}
To efficiently identify high-potential candidate responses at turn $t_j$, we perform a coarse-grained filtering step using an LLM as an automated judge. Given the dialogue history state $s_j = [t_i]_{i=1}^{j-1}$, the collector agent first generates $N$ candidate responses $\{a_j^{(1)}, a_j^{(2)}, \dots, a_j^{(N)}\}$ from its policy $\pi(a_j \mid s_j)$. An LLM judge then evaluates these candidates via a \textit{listwise comparison} prompt, assigning a quality score to each response based on three key dimensions: (1) \textbf{Respect \& Empathy}, (2) \textbf{Transparency} and (3) \textbf{Feasibility}. To mitigate positional bias, the evaluation is performed twice: once with the original order and once with the reversed order. Let $s_k^{\text{fwd}}$ and $s_k^{\text{rev}}$ denote the scores assigned to candidate $a_j^{(k)}$ in the forward and reverse evaluations, respectively. The combined score for each candidate is computed as:
\begin{equation}
    \mathcal{S}_k = s_k^{\text{fwd}} + s_k^{\text{rev}},
\end{equation}
where $\mathcal{S}_k$ represents the final quality score of candidate $a_j^{(k)}$. The top $K$ candidates are then selected as seed responses based on their combined scores.

\subsection{Fine-Grained Simulation}

Negotiation is a     multi-turn process requiring foresight—anticipating how each utterance shapes future dynamics. Inspired by experienced negotiators who refine tactics through practice, we use forward simulation to estimate long-term response impact via trajectory sampling, distinguishing effective strategies from superficial ones. To jointly optimize financial recovery and user experience, we define a \textit{Foresight Reward} over the full dialogue trajectory $\tau = [t_1, \dots, t_K]$, combining \textit{Agreement Quality} and \textit{Interaction Experience} rewards.


\paragraph{Agreement Quality Reward.}
The Agreement Quality Reward evaluates the financial favorability of the final repayment agreement achieved by the agent. Formally, it is defined as:
\begin{equation}
    R_{\text{aq}}(\tau) = f(\alpha),
\end{equation}
where $f(\cdot)$ is a domain-informed function that quantifies the financial favorability of the repayment agreement $\alpha$ to the creditor.
\paragraph{Interaction Experience Reward.}
The Interaction Experience Reward assesses the quality of the interaction from the debtor's perspective. We formally define the user satisfaction score $S$ as a composite measure derived from an LLM-based evaluation of empathy, respect, transparency, and communication ability. The formulation for this reward is as follows:
\begin{equation}
    R_{\text{ie}}(\tau) = \text{LLM}(\tau),
\end{equation}
By combining two critical components, the Foresight Reward is calculated as:
\begin{equation}
    R(\tau) = w_{\text{aq}} \cdot R_{\text{aq}}(\tau) + w_{\text{ie}} \cdot R_{\text{ie}}(\tau),
\end{equation}


\paragraph{Preference Optimization}
By combining coarse-grained filtering with fine-grained evaluation, we estimate the long-term impact of each candidate response through forward simulation using a persona-conditioned user agent. This yields a trajectory-level Foresight Reward that jointly captures task success and interaction quality. We then construct high-quality preference pairs by ranking responses based on their long-term outcomes and selecting the superior one as the preferred target. The agent is optimized via Direct Preference Optimization (DPO) \citep{dpo} to align its policy with these far-sighted preferences, reinforcing behaviors that lead to both effective agreements and positive user experiences.

\section{Experiment}

\begin{table*}[ht]
\vspace{-0.1in}
    \centering
    \vspace{-0.1in}
    \setlength{\tabcolsep}{3.5mm}{
    \resizebox{\textwidth}{!}{%
    \begin{tabular}{l|cccc|cc|ccc}
        \toprule
         \multirow{2}{*}{\textbf{Model}} & \multicolumn{4}{c}{\textbf{Negotiation Ability}} & \multicolumn{2}{c}{\textbf{Agreement Rationality}} & \multicolumn{3}{c}{\textbf{Interaction Experience}} \\
         
        \cmidrule(lr){2-5} \cmidrule(lr){6-7} \cmidrule(lr){8-10} 
        & \textbf{SR(\%)}$ \uparrow $ & \textbf{AT}$ \downarrow $ & \textbf{CR(\%)}$ \uparrow $ & \textbf{CE}$ \uparrow $ & \textbf{SA(\%)}$ \uparrow $ & \textbf{LS(\%)}$ \uparrow $ & \textbf{US}$ \uparrow $ & \textbf{ES}$ \uparrow $ & \textbf{CA}$ \uparrow $\\
        \midrule
        \rowcolor{gray!30} \multicolumn{10}{c}{\textbf{Close-source Large Language Models}} \\
        \midrule
        GPT-4o & \textbf{89.00}& \textbf{6.23}& \textbf{85.53}& \textbf{1.19}& \textbf{98.41}& 94.77& \underline{7.02}& 6.25& 8.36\\
        GPT-o1-mini & \underline{73.90}& \underline{7.79}& \underline{71.00}& \underline{1.15}& 93.50& 94.31& \textbf{7.12}& \underline{6.48}& 8.38\\
        DeepSeek-R1 & 51.93& 8.68& 50.81& 0.90& 96.58& \underline{95.77}& 5.75& 4.95& 7.75\\
        DeepSeek-V3 & 56.50& 8.26& 53.95& 0.75& 96.97& 94.65& 6.82& 6.28& \textbf{8.41}\\
        GLM-4.5 & 63.40& 8.38& 59.16& 0.86& 97.44& 95.52& 6.98& 6.45& \underline{8.40}\\
        Claude-4.0 & 43.54& 9.65& 42.38& 1.19& \underline{97.87}& 94.68& 5.65& 5.53& 7.74\\
        Gemini-2.5 & 49.10& 9.18& 48.29& 0.99& 93.87& \textbf{96.52}& 6.68& \textbf{6.79}& 8.32\\
        Kimi-K2 & 61.00& 8.40& 59.73& 0.98& 95.25& 94.43& 6.78& 6.14& 8.29\\
        Qwen3-235B & 71.80& 8.03& 66.86& 0.94& 97.45& 94.48& 6.91& 6.12& 8.33\\
        \midrule
        \rowcolor{gray!30} \multicolumn{10}{c}{\textbf{Open-source Large Language Models}} \\
        \midrule
        Llama-3-8B & 39.60& 9.48& 37.84& 0.61& 94.16& \underline{95.43}& 6.90& 6.42& 8.26\\
        Qwen3-8B & 73.20& 7.56& 69.39& 0.96& 94.88& 94.46& 7.17& 6.37& 8.39\\
        Qwen3-32B & 75.70& \underline{7.50}& \underline{73.32}& 1.06& 94.93& 94.79& \underline{7.24}& 6.40& \underline{8.43}\\
        QwQ-32B & 71.40& 7.83& 66.94& 0.92& \textbf{97.76}& 94.96& 6.76& 5.91& 8.23\\
        Llama-3-70B & \underline{76.20}& 8.03& 72.25& \textbf{1.21}& 92.38& 93.98& 7.11& 6.39& 8.40\\
        Qwen2.5-72B & 62.60& 8.36& 59.56& 0.82& \underline{95.65}& \textbf{95.49}& 7.00& \underline{6.52}& 8.40\\
        \midrule
        DebtGPT-8B & \textbf{84.10}& \textbf{7.22}& \textbf{79.54}& \underline{1.07}& 94.83& 94.83& \textbf{7.29}& \textbf{6.56}& \textbf{8.44}\\
        \bottomrule
    \end{tabular} }
    }
    \caption{\label{img:mainresult}The performances of advanced models as collectors (\underline{underlined} denotes the second-best performance)}
 \label{tab:mainresults}
     \vspace{-10pt}
\end{table*}

\subsection{Experimental Setup}
We comprehensively evaluated 16 models as collectors, fixing the debtor role to the advanced open-source model Qwen3-32B~\citep{qwen3}. Evaluation was conducted along three complementary axes: \textit{(1) Negotiation Ability}: measuring task success and efficiency via Success Rate (SR), Average Turn (AT), Collection Rate (CR), and Collection Efficiency (CE); \textit{(2) Agreement Rationality}: assessing economic sustainability for the debtor through Short-term Affordability (SA) and Long-term Sustainability (LS); and \textit{(3) Interaction Experience}: capturing user-centered quality across User Satisfaction (US), Emotion Support (ES), and Communication Ability (CA). Full metric definitions and computation details are provided in Appendix~\ref{appendix:experiment_detail}.

\subsection{Main Results}

\paragraph{\ding{172} Most existing models struggle in persona-enriched debt collection scenarios.}
The results in Table~\ref{tab:mainresults} highlight the limitations of \textit{current LLMs in handling the behavioral complexity inherent in debt collection}. Across the board, the majority of models achieve a success rate below 75\%, with several prominent systems such as Claude-4.0 and Llama-3-8B failing to secure agreements in more than half of the interactions. 
This widespread underperformance indicates a fundamental difficulty in adapting to the rich behavioral heterogeneity inherent in persona-enriched negotiation—spanning emotional volatility, cognitive constraints, and diverse linguistic styles.


\paragraph{\ding{173} Models frequently over-concede to secure agreements, undermining financial interests.}
Despite high Agreement Rationality scores—indicating that proposed repayment plans are affordable for debtors—most models exhibit low Collection Rate (CR) and Collection Efficiency (CE), reflecting a tendency to over-concede in order to secure agreements at the expense of financial recovery (e.g. offering large discounts, accepting upfront payments below the industry threshold). This behavior reflects a myopic strategy: \textit{rather than assessing the debtor’s true financial capacity or negotiating firm but fair terms, the model defaults to leniency as the path of least resistance}. Consequently, while agreements meet basic affordability constraints, they consistently underperform on financial recovery, exposing a fundamental failure to balance empathy with principle.

\begin{figure}[t]
\centering
\includegraphics[width=\linewidth, height=4cm]{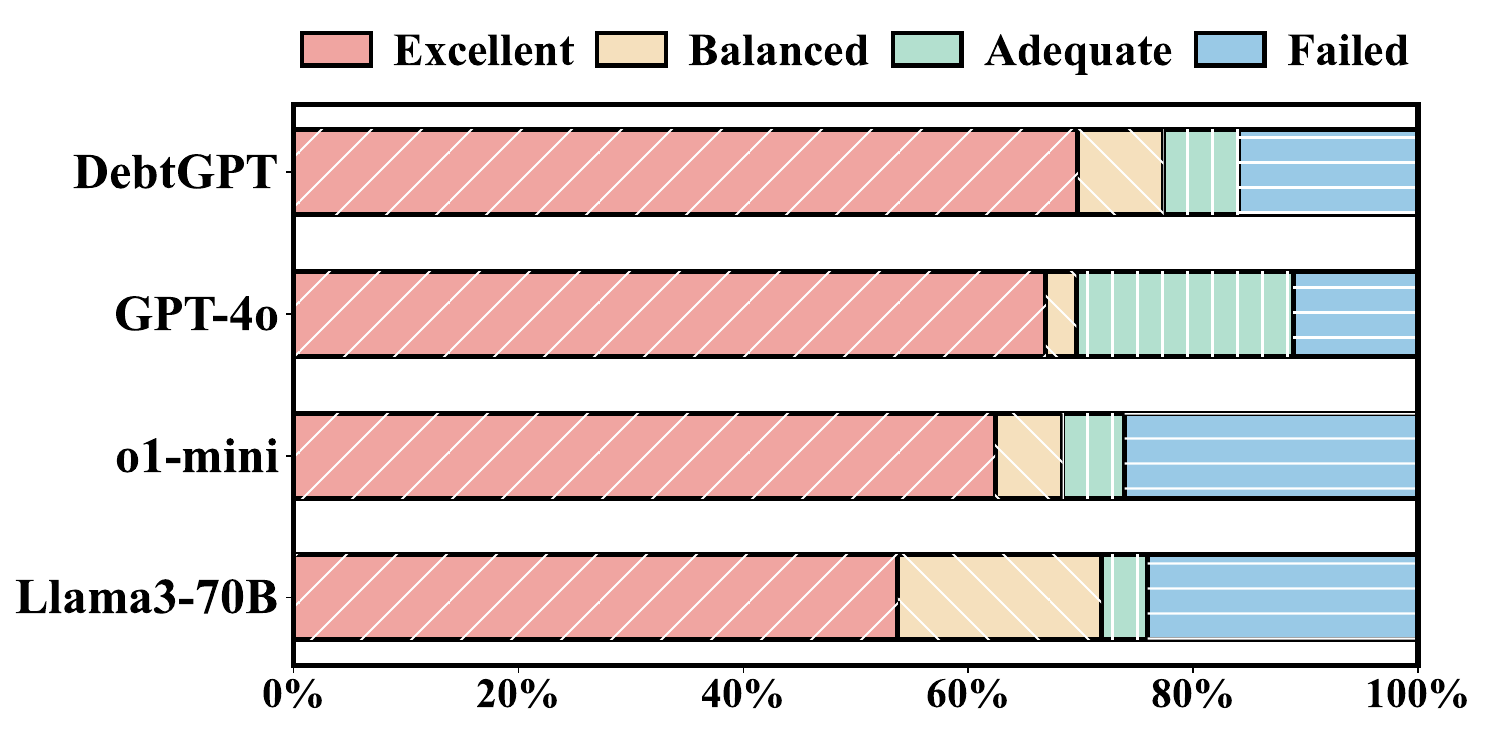}
\caption{Distribution of comprehensive performance ratings across models on the DebtBench benchmark. The overall score for each dialogue is computed by combining Agreement Quality and Interaction Experience.}
\label{fig:performance_distribution}
\end{figure}

\begin{table*}[ht]
\vspace{-0.1in}
    \centering
    \vspace{-0.1in}
    \renewcommand{\arraystretch}{1.3}
    \setlength{\tabcolsep}{3.5mm}{
    \resizebox{\textwidth}{!}{%
    \begin{tabular}{l|cccc|cc|ccc}
        \toprule
         \multirow{2}{*}{\textbf{Debtor Category}} & \multicolumn{4}{c}{\textbf{Negotiation Ability}} & \multicolumn{2}{c}{\textbf{Agreement Rationality}} & \multicolumn{3}{c}{\textbf{Interaction Experience}} \\
        \cmidrule(lr){2-5} \cmidrule(lr){6-7} \cmidrule(lr){8-10} 
        & \textbf{SR(\%)}$ \uparrow $ & \textbf{AT}$ \downarrow $ & \textbf{CR(\%)}$ \uparrow $ & \textbf{CE}$ \uparrow $ & \textbf{SA(\%)}$ \uparrow $ & \textbf{LS(\%)}$ \uparrow $ & \textbf{US}$ \uparrow $ & \textbf{ES}$ \uparrow $ & \textbf{CA}$ \uparrow $\\
        \midrule
        Confrontational (38\%) & 52.76$_{\textcolor{red}{\downarrow12.68}}$ & 8.95$_{\textcolor{red}{\downarrow0.81}}$ & 49.47$_{\textcolor{red}{\downarrow12.95}}$ & 0.74$_{\textcolor{red}{\downarrow0.24}}$ & \underline{97.11}$_{\textcolor{darkgreen}{\uparrow1.35}}$ & \underline{97.39}$_{\textcolor{darkgreen}{\uparrow2.45}}$ & 6.13$_{\textcolor{red}{\downarrow0.71}}$ & 5.63$_{\textcolor{red}{\downarrow0.62}}$ & 7.92$_{\textcolor{red}{\downarrow0.38}}$ \\
        Cooperative (6\%) & \textbf{86.06}$_{\textcolor{darkgreen}{\uparrow20.62}}$ & \textbf{6.20}$_{\textcolor{darkgreen}{\uparrow1.94}}$ & \textbf{83.36}$_{\textcolor{darkgreen}{\uparrow20.94}}$ & \textbf{1.84}$_{\textcolor{darkgreen}{\uparrow0.86}}$ & \textbf{98.57}$_{\textcolor{darkgreen}{\uparrow2.81}}$ & \textbf{99.61}$_{\textcolor{darkgreen}{\uparrow4.67}}$ & \textbf{7.76}$_{\textcolor{darkgreen}{\uparrow0.92}}$ & \textbf{6.99}$_{\textcolor{darkgreen}{\uparrow0.74}}$ & \textbf{8.78}$_{\textcolor{darkgreen}{\uparrow0.48}}$ \\
        Avoidant (31\%) & 
        65.14$_{\textcolor{red}{\downarrow0.30}}$ & 8.32$_{\textcolor{red}{\downarrow0.18}}$ & 62.78$_{\textcolor{darkgreen}{\uparrow0.35}}$ & 0.96$_{\textcolor{darkgreen}{\downarrow0.01}}$ & 96.21$_{\textcolor{darkgreen}{\uparrow0.46}}$ & 95.57$_{\textcolor{red}{\uparrow0.64}}$ & 6.98$_{\textcolor{darkgreen}{\uparrow0.14}}$ & 6.38$_{\textcolor{darkgreen}{\uparrow0.13}}$ & 8.38$_{\textcolor{darkgreen}{\uparrow0.08}}$ \\
        Helpless (25\%) & \underline{80.44}$_{\textcolor{darkgreen}{\uparrow14.99}}$ & \underline{7.14}$_{\textcolor{darkgreen}{\uparrow1.00}}$ & \underline{77.06}$_{\textcolor{darkgreen}{\uparrow14.63}}$ & \underline{1.16}$_{\textcolor{darkgreen}{\uparrow0.18}}$ & 93.27$_{\textcolor{red}{\downarrow2.48}}$ & 90.73$_{\textcolor{red}{\downarrow4.21}}$ & \underline{7.53}$_{\textcolor{darkgreen}{\uparrow0.69}}$ & \underline{6.89}$_{\textcolor{darkgreen}{\uparrow0.64}}$ & \underline{8.68}$_{\textcolor{darkgreen}{\uparrow0.38}}$ \\
        \midrule
        Overall (100\%) & 65.45 & 8.14 & 62.43 & 0.97 & 95.76 & 94.94 & 6.84 & 6.25 & 8.30 \\
        \bottomrule
    \end{tabular} }
   } \caption{\label{img:debtor_category}Comprehensive performance evaluation of collectors across various debtor categories. \textcolor{darkgreen}{Green} (\textcolor{red}{Red}) indicates the increased (decreased) performance compared to overall performance.}
 \label{tab:mainresults}
     \vspace{-10pt}
\end{table*}

\paragraph{\ding{174} Models fail to sustain positive interaction experiences during negotiation.}
Beyond task-level outcomes, the gap in Interaction Experience is equally pronounced: most models perform poorly across all three dimensions. Even the strongest closed-source models, such as GPT-4o (US: 7.02, ES: 6.25) and GPT-o1-mini (US: 7.12, ES: 6.48), achieve only moderate scores in user-centered metrics, while others like Claude-4.0 (US: 5.65, ES: 5.53) fall substantially short. Their interactions often come across as rigid, formulaic, or subtly coercive (e.g., emphasizing legal consequences without acknowledging distress), failing to adapt to the debtor’s emotional state or cognitive limitations. This erodes trust, triggers defensiveness, and leads to disengagement—ultimately compromising both user dignity and long-term collection efficacy.

\paragraph{\ding{175} Reasoning-specialized models underperform general-purpose counterparts in behaviorally complex negotiation.}
Contrary to expectations, models explicitly optimized for complex reasoning, like GPT-o1-mini, QwQ-32B and DeepSeek-R1 consistently lag behind general-purpose counterparts like GPT-4o and Qwen3-32B. For instance, GPT-o1-mini achieves only 73.9\% success rate and 71.0\% collection rate, markedly lower than GPT-4o’s 89.0\% and 85.5\%. Through deep analysis of their dialogue trajectories, we found that its failures stem not from logical inconsistency, but from \textit{a mismatch between its reasoning paradigm and the negotiation behavior of debt collection}. Specifically, the model often adheres rigidly to a formal, step-by-step inference process without adequately responding to the debtor’s different behaviors. For instance, when confronted with hostile or evasive debtors, it frequently persists with factual questioning rather than first de-escalating tension through empathy or ethical appeal, leading to premature negotiation breakdowns. This highlights that successful debt collection relies less on formal logic and more on dynamically adapting responses to the debtor’s real-time behavior—a capability that current reasoning-specialized models still lack.

\paragraph{\ding{176} DebtGPT achieves the best balance between financial recovery and user experience.}
As shown in Figure~\ref{fig:performance_distribution}, we jointly evaluate Agreement Quality and Interaction Experience to analyze the dialogue-level performance of top models: GPT-4o, GPT-o1-mini, Llama-3-70B, and DebtGPT. Among all evaluated agents, DebtGPT demonstrates the most consistent ability to secure agreements that are both economically effective and user-respectful. This results from our Coarse-to-Fine Preference Optimization (CFPO) framework, which explicitly optimizes for maximizing recovery while preserving user dignity. By combining filtering with targeted forward simulation, CFPO enables far-sighted, behaviorally adaptive strategies that align institutional goals with human-centered interaction.

\paragraph{\ding{177} Collector performance varies significantly across debtor behavioral categories.}
We further conduct a fine-grained evaluation by categorizing debtors into four distinct behavioral types based on their profiles. As shown in Table~\ref{tab:mainresults}, the results reveal that negotiation outcomes are highly sensitive to the debtor’s behavioral profile. Cooperative debtors yield the strongest performance, far exceeding the overall average. In contrast, confrontational debtors pose the greatest challenge: SR drops to 52.76\% and CR to 49.47\%, with notably lower user satisfaction (US: 6.13) and emotion support (ES: 5.63), reflecting the difficulty of maintaining rapport under hostility. These findings underscore a critical insight: effective debt collection requires not just strategic reasoning, but dynamic adaptation to the debtor’s behavioral type—a capability that remains underdeveloped in most existing agents.

\section{Analysis LLM-as-Judge}
Following prior work on LLM-as-a-judge reliability \citep{sotopia}, we conducted a human correlation study to validate the reliability of our LLM-based evaluation for Interaction Experience. We randomly selected 200 dialogue samples from our test set and had them scored independently by five human annotators and our LLM judge (DeepSeek-V3). The $\kappa$ score between human annotators is 0.566, which indicates fair to moderate inter-annotator agreement.

\begin{figure}[H]
\centering
\includegraphics[width=\linewidth,height=3.5cm]{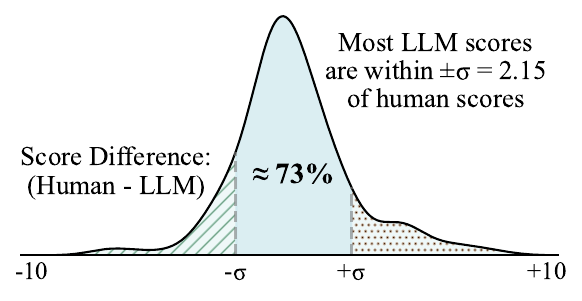}
\caption{Distribution of Human–LLM Differences in Interaction Experience Scores.}
\label{fig:score_compariso}
\end{figure}

As shown in Figure \ref{fig:score_compariso},  the majority (nearly 73\%) of LLM scores concentrate around the human scores within a standard deviation, indicating strong alignment. Despite known evaluator biases \citep{positionalbias, native}, our results suggest that the LLM can serve as a reliable proxy for human judgment when guided by a carefully designed rubric. See Appendix~\ref{app:llm_human} for details.

\section{Conclusion}
In this paper, we extend negotiation dialogue research to the high-stakes and behaviorally complex domain of debt collection. We propose DebtBench, the first public, persona-enriched benchmark that captures the rich behavioral heterogeneity observed in practice. Our evaluation of 16 state-of-the-art LLMs reveals that most still struggle in this setting, often over-conceding for superficial agreements or failing to adapt to user behavior. Moreover, we develop DebtGPT, a negotiation agent trained to jointly optimize long-term financial recovery and human-centered interaction experience, effectively harmonize institutional efficacy with positive interaction experience. This work provides valuable insights and a responsible testbed for advancing behaviorally grounded, empathetic, and socially aware negotiation systems in high-stakes domains.

\section*{Limitations}
We discover some limitations during benchmark construction. (1) While DebtBench emphasizes behavioral heterogeneity, it models financial attributes (e.g., assets, daily income) as static values, abstracting away real-world dynamics such as income volatility or temporary default—factors critical to long-term repayment feasibility. (2) Despite our efforts to generate behaviorally diverse and realistic dialogues through persona-driven simulation, the model-generated utterances may still diverge from the linguistic and strategic patterns of actual debt collection conversations, potentially limiting domain-specific applicability. (3) DebtBench is constructed in collaboration with a single fintech firm within a specific setting. As debt collection practices are also shaped by cultural norms and legal regulations, the behavioral patterns in DebtBench may not fully generalize to other regions or jurisdictions. Nevertheless, given the extreme privacy sensitivity and scarcity of real-world data in this high-stakes setting, we believe DebtBench provides a valuable, ethically sound testbed for advancing behavior-aware negotiation systems.

\section*{Ethical Considerations}
Our work strictly adheres to privacy and ethical standards. No real user data is released; all source dialogues were obtained under strict confidentiality agreements with a major internet financial institution and approved by its internal compliance board. We fully anonymize sensitive information—replacing names with pseudonyms, synthesizing financial attributes, and generalizing delinquency reasons into high-level categories (e.g., “job loss”)—ensuring no personally identifiable or traceable details remain. Every synthetic persona and dialogue in DebtBench underwent manual verification to guarantee privacy preservation. DebtBench is a simulation-based benchmark designed solely for academic research, and the proposed DebtGPT agent operates in a controlled environment without real-world deployment. Any future operational use would require rigorous human-in-the-loop validation, regulatory review, and explicit safeguards to protect vulnerable individuals.

We conducted two human evaluation studies: (1) assessing the realism and persona consistency of DebtBench-generated dialogues, and (2) validating our LLM-as-Judge via a human correlation study. A total of 20 annotators participated, all of whom have professional experience in financial services. All participants provided informed consent and signed a disclaimer acknowledging the purpose of the study, data usage, and voluntary nature of participation. The dialogues presented for annotation contained no personally identifiable or sensitive financial information, as all real user data in the original conversations had been anonymized. The tasks posed minimal risk, involving only reading and rating pre-existing dialogues.

\section*{Potential Risks}
While our work aims to advance human-centered dialogue systems in high-stakes financial negotiation, we acknowledge several potential risks inherent to this domain.

First, debt collection involves highly sensitive personal and financial data, raising legitimate privacy and misuse concerns. To mitigate these risks, DebtBench contains no real user information—all personas and dialogues are synthetically generated to preserve behavioral and statistical patterns of real interactions only at the distributional level, with no traceable links to individuals. Names, financial details, and life events are replaced by pseudonyms, modeled surrogates, or abstract categories (e.g., “job loss”), and every entry is manually verified for privacy compliance. A second concern is the potential misuse of DebtBench to train adversarial “anti-collection” agents that evade legitimate debt resolution. While this risk exists for any negotiation benchmark, our data is explicitly designed to model good-faith, cooperative negotiation under financial hardship—not evasion tactics—and will be released under a research-only license prohibiting operational or adversarial use.

Ultimately, our goal is establish a privacy-preserving, ethically grounded testbed for studying behaviorally adaptive dialogue systems in high-stakes, human-centered scenarios—domains long excluded from NLP research due to data sensitivity.

\section*{Acknowledgments}
This paper was mainly supported by the NSFC under Grants (No. 62402424). This work was also supported by Ant Group.


\bibliography{custom}

\appendix

\section{Debt Collection Negotiation}
\label{appendix:debttalk_detail}

\subsection{Task Formulation}
We formulate the debt collection negotiation task as a Markov Decision Process (MDP). Each turn $t_j = (u_j, r_j)$ is represented as a pair consisting of the collector's utterance $u_j$ followed by the debtor's response $r_j$, where $K$ denotes the total number of turns in the conversation. At each turn $t_j$, the collector agent observes the dialogue history state $s_j = [t_i]_{i=1}^{j-1}$ and selects an action $a_j \in \mathcal{A}$, where $\mathcal{A}$ is a set of candidate negotiation strategies defined by domain experts. The interaction then proceeds with the debtor's response, and this process repeats until the repayment agreement is achieved or the maximum number of turns $T$ is reached.

In real-world debt collection scenarios, the agreement $\alpha$ is characterized by four key dimensions: \textit{Discount Ratio}, \textit{Immediate Payment Ratio}, \textit{Immediate Payment Time}, and \textit{Installment Periods}, with detailed definitions provided in Appendix~\ref{appendix:negotiation_dimension}. The objective is to learn a policy $\pi(a_j \mid s_j)$ that generates efficient negotiation trajectories leading to a \textbf{mutually acceptable repayment agreement}, while preserving a \textbf{positive user experience}. Formally, we define the objective as:
\begin{equation}
    \max_{\pi}  \mathbb{E}_{\tau \sim \pi} \left[ R(\tau) \right]
\end{equation}
where $\tau = [t_1, t_2, \dots, t_K]$ denotes a complete dialogue trajectory, and $R(\tau)$ is a reward function that incorporates task success—measured by the agreement dimensions—as well as user experience factors, such as the degree of respect and empathy conveyed by the collector's language.

\subsection{Four Dimensions of Debt Collection Negotiation}
\label{appendix:negotiation_dimension}
In real-world debt collection scenarios, the negotiation process centers around four key agreement dimensions, which jointly determine the economic outcome and repayment feasibility. These dimensions are not only standard practice in financial institutions but also directly shape the strategic decisions of both collectors and debtors. Formally, a repayment agreement $\alpha$ is characterized by:
\begin{itemize}[leftmargin=2.5mm,parsep=2pt]

\item \textbf{Discount Ratio (disc\_ratio)}: The percentage of the total debt that may be waived by the creditor to ease the debtor’s repayment burden. In our setting, allowable discount levels are discrete: \{0\%, 5\%, 10\%, 15\%, 20\%, 25\%, 30\%\}. Discounts are only justified when the debtor demonstrates genuine financial hardship or is in severe distress. Absent such conditions, agents are encouraged to avoid offering discounts to preserve institutional interests.

\item \textbf{Immediate Payment Ratio (pmt\_ratio)}: The proportion of the total debt that must be paid upfront upon agreement. Valid values range from 5\% to 50\% in 5\% increments. To ensure meaningful commitment and reduce default risk, a minimum threshold of 25\% is typically enforced, though flexibility is permitted based on the debtor’s liquidity constraints.

\item \textbf{Immediate Payment Time (pmt\_days)}: The number of days granted to the debtor to complete the immediate payment, with a maximum of 14 days. While same-day payment is ideal, a grace period of up to 7 days is standard; extensions beyond 7 days are only granted when the debtor provides credible evidence of temporary cash flow constraints.

\item \textbf{Installment Periods (inst\_prds)}: The number of months over which the remaining debt (after discount and immediate payment) will be repaid. Options include \{3, 6, 9, 12, 18, 24\} months, each associated with a predefined interest rate schedule. Shorter terms (e.g., 3 or 6 months) are preferred to minimize credit risk and accelerate recovery, and should be prioritized unless the debtor faces exceptional hardship.
\end{itemize}
These four dimensions constitute the core negotiation space in actual debt collection operations and serve as the basis for both task success evaluation and strategic planning in our framework.

\section{Details of DebtBench}

\subsection{Persona in DebtBench}
Due to the wide variation in debtors' financial conditions, personal backgrounds, and psychological traits in real-world debt collection, it is imperative for collector agents to generate personalized and adaptive negotiation strategies tailored to diverse users. To support such capabilities, we introduce the \textbf{DebtBench}, a synthetic yet realistic dataset built upon real interaction data. We construct rich, multidimensional personas and use them to generate diverse and emotionally expressive dialogues through role-play simulation between collector and debtor agents, resulting in the first debt collection dialogue dataset comprehensively grounded in persona-enriched user modeling.

\paragraph{Background Information} This dimension includes personal attributes, debt-related information, and financial status. To preserve real-world data characteristics while ensuring privacy, we fit a multivariate Gaussian model to 11,000 real debt records from a leading financial institution and generate 100,000 synthetic profiles. Each real user is matched to the most similar synthetic profile using k-nearest neighbors (k-NN) based on demographic and debt-related features. The debtor's financial status includes overdue reason and economic condition. The overdue reason categories were defined by financial experts and we use an LLM to assign each user to the most plausible category based on their real dialogue history. Economic indicators are generated by a second LLM conditioned on the full user profile to ensure contextual consistency. We compare the distribution of real and synthetic user background information in Figure ~\ref{fig:distribution_compariso}, showing strong alignment in PCA space.

\paragraph{Personality Traits} Debt collection dialogues are rich in emotional expression, with user responses shaped by personality traits like temperament and resilience. To capture this behavioral diversity, we construct a comprehensive personality dimension, capturing five key aspects: character, MBTI type, emotion, emotional resilience, and linguistic style. Character reflects the debtor’s dispositional traits (e.g., responsible, impatient) that shape their attitude and communication tone. MBTI personality type provides a structured psychological framework for predicting behavioral tendencies. Emotion captures dynamic affective states (e.g., anxious, defensive) that evolve across dialogue turns, while emotional resilience represents the individual’s stable ability to cope with stress in high-pressure interactions. Linguistic style describes verbal preferences (e.g., evasive, confrontational) that influence expression patterns. All traits are inferred by LLMs through joint analysis of the user’s background and real collector-debtor dialogue history. This ensures behavioral consistency and contextual alignment, enabling realistic and profile-grounded role-playing simulation.

\paragraph{Cognitive Attributes} Through analysis of real collector-debtor dialogues, we observe that debtors often exhibit cognitive deficits—such as lack of legal awareness or financial knowledge—that lead to confusion or non-cooperative behavior in conversation. However, LLMs inherently possess strong domain knowledge, making it difficult to reproduce low-awareness but authentic behavior. To bridge this gap between model capability and user reality, we introduce a cognitive dimension comprising four attributes: legal awareness, financial literacy, responsibility, and credit awareness. These attributes are inferred by LLMs through joint analysis of the user’s background and dialogue history, enabling behaviorally grounded and role-consistent simulations. This design ensures that the debtor model reflects the individual’s actual understanding rather than the model’s internal knowledge, thereby supporting more realistic role-playing interactions.

\paragraph{Life-grounded Scenario} To ensure that each user is not just a collection of attributes but a coherent individual, we generate a life-grounded scenario for every persona. This scenario is a concise narrative—such as job loss due to company downsizing, medical debt from a family illness, or financial strain after a failed investment—that describes the user’s current life situation, explains the origin of their debt, shapes their behavioral tendencies in negotiation, and is generated by an LLM based on the full user profile. By aligning the narrative with the persona, we ensure behaviorally consistent and life-grounded role-playing simulation, where the debtor’s responses are grounded in a plausible and emotionally resonant life context.

\subsection{Detailed Strategy Extraction Pipeline}
To provide greater transparency on the construction of the strategy taxonomy in Tables \ref{tab:collector_strat}--\ref{tab:debtor_strat}, we describe here the full strategy extraction pipeline and the measures taken to reduce potential bias in expert refinement. As briefly introduced in the main paper, our strategy extraction process starts from real collector–debtor conversations, followed by strategy clustering and expert consolidation into the final taxonomy.

\subsubsection{Data-anchored strategy extraction}
Our strategy set was derived directly from 1,000 real-world dialogues between experienced human collectors and debtors. This design ensures that the extracted strategies are grounded in authentic interaction patterns observed in practice, rather than being manually invented from abstract assumptions. In other words, the taxonomy is intended to be descriptive of recurring real-world behaviors, instead of being a purely theoretical categorization.

Concretely, we first prompted a strong LLM to identify salient strategy–utterance pairs from the real dialogue corpus. Each pair consists of a short strategy description and its supporting utterance span. We then embedded these pairs into a semantic space and applied HDBSCAN clustering to group behaviorally similar strategies. This step helped reduce redundancy, surface recurring negotiation patterns, and provide a data-driven starting point for subsequent human refinement.

\subsubsection{Multi-expert refinement and validation}
After clustering, the preliminary strategy inventory was reviewed through an iterative multi-expert refinement process. For each cluster, representative utterances were manually inspected to determine whether the cluster corresponded to a coherent, interpretable, and reusable negotiation behavior. Ambiguous or overly broad clusters were split, merged, or discarded, and the resulting strategy definitions were repeatedly validated against authentic conversation patterns.

This process was designed to reduce individual annotator bias in two ways. First, expert decisions were anchored to clusters induced from real dialogue data rather than from top-down manual definitions. Second, strategy consolidation was performed through multi-expert discussion and consensus, instead of relying on a single reviewer. In this sense, the final taxonomy reflects not only data-driven behavioral regularities, but also agreement across multiple reviewers regarding whether a candidate strategy is semantically distinct and practically meaningful.

\subsection{Behavior Refinement Details}

To ensure that the synthesized dialogues faithfully reflect the assigned debtor personas, we introduce a behavior refinement stage after persona construction. The goal of this step is to improve the alignment between persona-driven dialogue behaviors and the corresponding persona attributes, particularly along the three dimensions emphasized in the main paper: emotional consistency, cognitive plausibility, and linguistic style coherence.

Specifically, given an initial persona profile and a simulated dialogue trajectory, we prompt an LLM to act as a behavior consistency evaluator. The model is asked to inspect whether the behaviors expressed in the dialogue are compatible with the assigned persona. If inconsistencies are identified, the model is further instructed to update the specific persona fields that conflict with the dialogue evidence, while keeping the remaining fields unchanged. This refinement process is repeated iteratively until the persona is behaviorally aligned with the simulated dialogue or the maximum refinement step is reached. The evaluation focuses on the following three aspects:

\begin{itemize}
    \item \textbf{Emotional Consistency:} whether the debtor's emotional tendencies and personality traits reflected in the dialogue are consistent with those specified in the persona. For example, a highly defensive debtor should not consistently exhibit an overly calm or cooperative tone without sufficient conversational justification.
    \item \textbf{Cognitive Plausibility:} whether the behaviors expressed in the dialogue are compatible with the debtor's cognitive attributes, such as legal awareness, financial literacy, sense of responsibility, and understanding of credit consequences. For instance, a debtor with limited legal or financial knowledge should not repeatedly produce highly technical explanations.
    \item \textbf{Linguistic Style Coherence:} whether the wording, tone, and expression style observed in the dialogue are consistent with the persona, such as being evasive, fragmented, cautious, or confrontational.
\end{itemize}

In practice, the refinement model is prompted to produce both an analysis and, when necessary, updates to the inconsistent persona fields rather than rewriting the dialogue itself.

\subsection{Ablation Study on Prompt Design}

We conduct an ablation study to validate the contribution of persona profiles and negotiation strategies to dialogue generation. As shown in Table~\ref{tab:different_prompt}, incorporating either component improves human-likeness and realism over a naive baseline; combining both yields the strongest performance across all dimensions.

\begin{table}[!ht]
\centering
\small
\setlength{\tabcolsep}{5pt}
\renewcommand{\arraystretch}{1.1}
\label{tab:different_prompt}
\begin{adjustbox}{width=\columnwidth, center}
\begin{tabularx}{\linewidth}{
    >{\centering\arraybackslash}X |
    c c c |
    c c c
}
\toprule
\multirow{2}{*}{\textbf{Prompt Type}} & \multicolumn{3}{c}{\textbf{Human.}} & \multicolumn{3}{c}{\textbf{Real.}}\\
\cmidrule(lr){2-4} \cmidrule(lr){5-7}
& \textbf{W} & \textbf{T} & \textbf{L} & \textbf{W} & \textbf{T} & \textbf{L}\\
\midrule
\textbf{Naive} & - & - & - & - & - & - \\
\textbf{Prompt w/ Stra.} & 397 & 530 & 73 & 483 & 317 & 200 \\
\textbf{Prompt w/ Prof.} & 540 & 361 & 99 & 531 & 370 & 99 \\
\textbf{Prompt w/ Both.} & \textbf{693} & 260 & 47 & \textbf{712} & 271 & 17 \\
\bottomrule
\end{tabularx}
\end{adjustbox}
\caption{Evaluation of different prompts based human-likeness and realism. Scores are presented for three evaluation measures: Win (W), Tie (T), and Lose (L).}
\label{tab:different_prompt}
\end{table}

\section{Detail of Experiment}
\label{appendix:experiment_detail}

\subsection{Baselines}
We comprehensively evaluated 16 models as collector, with parameters ranging from 7B to 70B: 1) \textbf{Close-source language model}, including GPT-4o \citep{gpt}, DeepSeek-R1 \citep{deepseekr1}, DeepSeek-V3 \citep{deepseek-v3}, Claude Sonnet 4 \citep{claude-sonnet4}, Gemini2.5-pro \citep{gemini}, GLM-4.5 \citep{glm} and Kimi-K2-Instruct \citep{kimi}. 2) \textbf{Open-source language model}, including Qwen-series \citep{qwen3, qwen2.5} and Llama3 \citep{llama3}. To ensure dialogue quality and reproducibility, we fixed the debtor role to the advanced open-source model Qwen3-32B.

\subsection{Implementation Details}
We deploy open-source models on 8 H20 GPUs using the vLLM \citep{vLLM}, while closed-source
models are accessed through official APIs in accordance with their documentation. Additionally, we set the model’s temperature to 0 to ensure deterministic outputs (Despite setting temperature to 0, closed-source models accessed through APIs may still produce slight variations due to their internal non-deterministic mechanisms). The maximum output length is set to 1,024 tokens for non-reasoning models and 4,096 tokens for reasoning-specialized models (e.g., GPT-o1-mini, QwQ-32B), to accommodate their extended reasoning traces. Specific model hyperparameters and version details can be found in Table~\ref{tab:model-hyperparams}. Our DebtGPT are based on Qwen3-8B \citep{qwen3} with Lora fine-tuning \citep{lora} using Llama-Factory framework \citep{llamafactory}.

\subsection{Metrics.}
As previously mentioned, the goal of debt collection is to generate efficient negotiation trajectories leading to a mutually acceptable repayment agreement, while preserving a positive user experience. Accordingly, we design evaluation metrics along three complementary axes: (1) \textbf{Negotiation Ability}: Quantify the economic outcome and operational efficiency, including Success Rate (SR), Average Turn (AT), Collection Rate (CR), and Collection Efficiency (CE). (2) \textbf{Agreement Rationality}: Evaluate whether the negotiated repayment plan is sustainable for the debtor, considering both Short-term Affordability (SA) and Long-term Sustainability (LS). (3) \textbf{Interaction Experience}: Assess the quality of the interaction across three dimensions: User Satisfaction (US), Emotion Support Ability (ES), and Communication Ability (CA).
Details of metric computation and evaluation protocols are provided in Appendix \ref{appendix:metrics}.

\label{appendix:metrics}

\subsubsection{Negotiation Ability}
Negotiation Ability evaluates the agent’s operational effectiveness in achieving the core objectives of debt collection: securing a repayment agreement efficiently while maximizing financial recovery. This dimension captures both the \textit{success} and \textit{efficiency} of the negotiation process:

\paragraph{Success Rate (SR)}
Success Rate measures the proportion of dialogues in which the collector agent successfully reaches a mutually acceptable repayment agreement with the debtor within a pre-defined maximum number of turns $T_{\max}$. Formally, given a set of $N$ evaluation dialogues $\{\tau^{(i)}\}_{i=1}^N$, let $\mathbb{I}_{\text{succ}}(\tau^{(i)})$ be an indicator function that equals 1 if dialogue $\tau^{(i)}$ terminates with a valid agreement (i.e., all four agreement dimensions—discount ratio, immediate payment ratio, payment time, and installment periods—are fully specified and fall within allowable ranges) before or at turn $T_{\max}$, and 0 otherwise. Then:
\begin{equation}
    \text{SR} = \frac{1}{N} \sum_{i=1}^{N} \mathbb{I}_{\text{succ}}(\tau^{(i)}),
\end{equation}

\paragraph{Average Turn (AT)}
Average Turn reflects the overall dialogue efficiency by measuring the average number of turns across \textit{all} evaluation dialogues, regardless of whether an agreement was reached. This includes both successful negotiations and those that terminated without agreement (e.g., due to user disengagement or timeout). Let $T^{(i)}$ denote the total number of turns in the $i$-th dialogue (capped at the maximum allowed turns $T_{\max}$ if no agreement is reached). Formally:
\begin{equation}
    \text{AT} = \frac{1}{N} \sum_{i=1}^{N} T^{(i)},
\end{equation}
A lower AT indicates more concise and efficient communication across the entire dialogue.

\paragraph{Collection Rate (CR)}
Collection Rate quantifies the financial effectiveness of the negotiation by measuring the portion of the original debt amount recovered by the creditor. Following standard practice in credit risk literature, for a successful agreement $\alpha^{(i)}$ with discount ratio $dr^{(i)}$, the recovery ratio is defined as $r_i = 1 - dr^{(i)}$. For unsuccessful dialogues, the recovery ratio is considered 0. The Collection Rate is then the mean recovery ratio across all test samples:
\begin{equation}
    \text{CR} = \frac{1}{N} \sum_{i=1}^{N} \left[ (1 - dr^{(i)}) \cdot \mathbb{I}_{\text{succ}}(\tau^{(i)}) \right],
\end{equation}
where $d^{(i)}$ denotes the discount ratio agreed upon in the $i$-th dialogue, and $\mathbb{I}_{\text{success}}(\tau^{(i)})$ is an indicator function that equals 1 if the dialogue ends with a valid agreement, and 0 otherwise.
\paragraph{Collection Efficiency (CE)}
Collection Efficiency measures the expected daily recovery rate of the debt, reflecting how quickly the creditor can recoup funds under the agreed repayment plan. It is computed only over successful dialogues, as no recovery occurs otherwise. Formally:
\begin{equation}
\begin{split}
    \text{CE} = \sum_{i=1}^{N} \Biggl[ &(1 - dr^{(i)}) \left( \frac{pr^{(i)}}{pd^{(i)}} + \frac{1 - pr^{(i)}}{ip^{(i)} \cdot 30} \right) \\
    &\cdot \mathbb{I}_{\text{succ}}(\tau^{(i)}) \Biggr],
\end{split}
\end{equation}
where $dr^{(i)}$, $pr^{(i)}$, $pd^{(i)}$, and $ip^{(i)}$ represent the discount ratio, immediate payment ratio, immediate payment days, and installment period in the $i$-th dialogue respectively.

\subsubsection{Agreement Rationality}
Research has shown that the longer a debtor remains in a state of severe financial distress, the higher their likelihood of defaulting on the loan~\citep{determinants}. Therefore, Agreement Rationality evaluates whether the negotiated repayment plan is economically sustainable for the debtor, considering both short-term liquidity and long-term income capacity. It ensures that the agreement does not impose excessive financial stress that could lead to default or hardship, which is quantified by two complementary metrics:

\paragraph{Short-term Affordability (SA)}
SA measures whether the debtor’s short-term financial buffer is sufficient to cover the immediate payment obligation. Let $A^{(i)}$ denote the debtor’s current assets, $I^{(i)}$ the daily income, $D^{(i)}$ the original debt amount. With a safety margin coefficient $w_{\text{short}} \in (0,1]$, the Short-term Affordability Index is defined as:
\begin{equation}
    \text{SA}^{(i)} = \frac{w_{\text{short}} \cdot \left( A^{(i)} + I^{(i)} \cdot pd^{(i)} \right)}{D^{(i)} \cdot (1 - dr^{(i)}) \cdot pr^{(i)}},
\end{equation}

\paragraph{Long-term Sustainability (LS)}
LS evaluates whether the debtor’s monthly income is sufficient to sustain the installment payments. With a long-term safety coefficient $w_{\text{long}}\in (0,1]$, the Long-term Sustainability Index is:
\begin{equation}
    \text{LS}^{(i)} = \frac{w_{\text{long}} \cdot I^{(i)} \cdot 30 \cdot ip^{(i)}}{D^{(i)} \cdot (1 - dr^{(i)}) \cdot (1 - pr^{(i)})},
\end{equation}

In our implementation, we set $w_{\text{short}} = 0.85$ and $w_{\text{long}} = 0.95$ based on domain guidelines from financial risk management practices. Both SA and LS are computed only for successful dialogues. The final SA and LS scores (reported in \%) are defined as:
\begin{align}
    \text{SA} &= \frac{\sum_{i=1}^{N} \mathbb{I}_{\text{succ}}(\tau^{(i)}) \cdot \mathbb{I}\left( \text{SA}^{(i)} \geq 1.0 \right)}{N_{\text{succ}}}, \\
    \text{LS} &= \frac{\sum_{i=1}^{N} \mathbb{I}_{\text{succ}}(\tau^{(i)}) \cdot \mathbb{I}\left( \text{LS}^{(i)} \geq 1.0 \right)}{N_{\text{succ}}},
\end{align}
where $N_{\text{succ}} = \sum_{i=1}^{N} \mathbb{I}_{\text{succ}}(\tau^{(i)})$. Higher scores indicate more economically rational and user-resilient agreements.

\subsubsection{Interation Experience}
Recent studies have demonstrated that large language models (LLMs) can serve as reliable and scalable evaluators of dialogue quality, achieving strong alignment with human judgments in dimensions such as empathy, coherence, and user-centeredness \citep{g-eval}. Building on this foundation, we leverage an LLM-based evaluator to assess the interaction experience from the debtor’s perspective along three key axes:

\begin{itemize}[leftmargin=2.5mm,parsep=2pt]
\item \textbf{User Satisfaction (US)}: Measures the overall perceived quality of the interaction, including whether the debtor feels respected, understood, and fairly treated. High satisfaction indicates that the collector’s strategy successfully balances institutional objectives with human dignity.

\item \textbf{Emotion Support Ability (ES)}: Evaluates the agent’s capacity to recognize, validate, and appropriately respond to the debtor’s emotional states (e.g., anxiety, defensiveness, or hopelessness). This includes avoiding dismissive or coercive language and offering empathetic reassurance when appropriate.

\item \textbf{Communication Ability (CA)}: Assesses the clarity, transparency, and adaptiveness of the collector’s language. A high score reflects concise explanations of terms, avoidance of jargon, and responsiveness to the debtor’s cognitive level and linguistic style (e.g., simplifying legal consequences for users with low financial literacy).
\end{itemize}

For each completed dialogue trajectory $\tau$ , we prompt a strong LLM (DeepSeek-V3) with a structured evaluation template that provides the full conversation history and the debtor’s persona profile. The evaluator assigns a score from 1 to 10 for each dimension, grounded in behavioral anchors derived from domain guidelines. Final scores are averaged across all test dialogues. This approach ensures that interaction quality is assessed not in isolation, but in context—accounting for the debtor’s unique behavioral profile and emotional trajectory.

\begin{table*}[ht]
    \centering
    \setlength{\tabcolsep}{3.5mm}{
    \resizebox{\textwidth}{!}{%
    \begin{tabular}{l|cccc|cc|ccc}
        \toprule
         \multirow{2}{*}{\textbf{Model}} & \multicolumn{4}{c}{\textbf{Negotiation Ability}} & \multicolumn{2}{c}{\textbf{Agreement Rationality}} & \multicolumn{3}{c}{\textbf{Interaction Experience}} \\
         
        \cmidrule(lr){2-5} \cmidrule(lr){6-7} \cmidrule(lr){8-10} 
        & \textbf{SR(\%)}$ \uparrow $ & \textbf{AT}$ \downarrow $ & \textbf{CR(\%)}$ \uparrow $ & \textbf{CE}$ \uparrow $ & \textbf{SA(\%)}$ \uparrow $ & \textbf{LS(\%)}$ \uparrow $ & \textbf{US}$ \uparrow $ & \textbf{ES}$ \uparrow $ & \textbf{CA}$ \uparrow $\\
        \midrule
       DebtGPT (Full)   & 84.10 & 7.22 & 79.54 & 1.07 & 94.83 & 94.83 & 7.29 & 6.56 & 8.44 \\
        w/o Simulation   & 77.32 & 7.35 & 65.43 & 0.94 & 93.21 & 92.76 & 7.12 & 6.12 & 8.30 \\
        w/o Filter       & 83.61 & 7.38 & 79.41 & 1.01 & 94.81 & 94.12 & 7.21 & 6.38 & 8.47 \\
        Qwen3-8B         & 73.20 & 7.56 & 69.39 & 0.96 & 94.88 & 94.46 & 7.17 & 6.37 & 8.39 \\
        \bottomrule
    \end{tabular} }
    }
    \caption{Ablation study results.}
 \label{tab:ablation}
\end{table*}

\section{Ablation Study of DebtGPT}
To further validate the effectiveness of our Coarse-to-Fine Preference Optimization (CFPO) framework, we conduct ablation studies by removing each core component:

\begin{itemize}[left=0pt]
    \item \textbf{w/o Filter}: Directly applying forward simulation to all candidate responses without coarse-grained filtering;
    \item \textbf{w/o Simulation}: Using only the coarse-grained LLM judge scores for preference ranking, without fine-grained foresight reward estimation.
\end{itemize}

As shown in Table~\ref{tab:ablation}, both variants underperform the full DebtGPT model. The \textit{w/o Filter} variant suffers from high computational cost and noisy reward signals, leading to unstable training and suboptimal policy convergence. More critically, the \textit{w/o Simulation} variant reverts to myopic behavior—offering excessive concessions to maximize short-term likability—resulting in significantly lower Collection Rate, despite decent user satisfaction. These results confirm that both filtering and simulation are essential: filtering ensures efficiency

\section{Case Study}

\subsection{Behavioral Diversity in Persona-Driven Negotiation}
\begin{figure*}[t]
\centering
\includegraphics[width=\linewidth, height = 9cm]{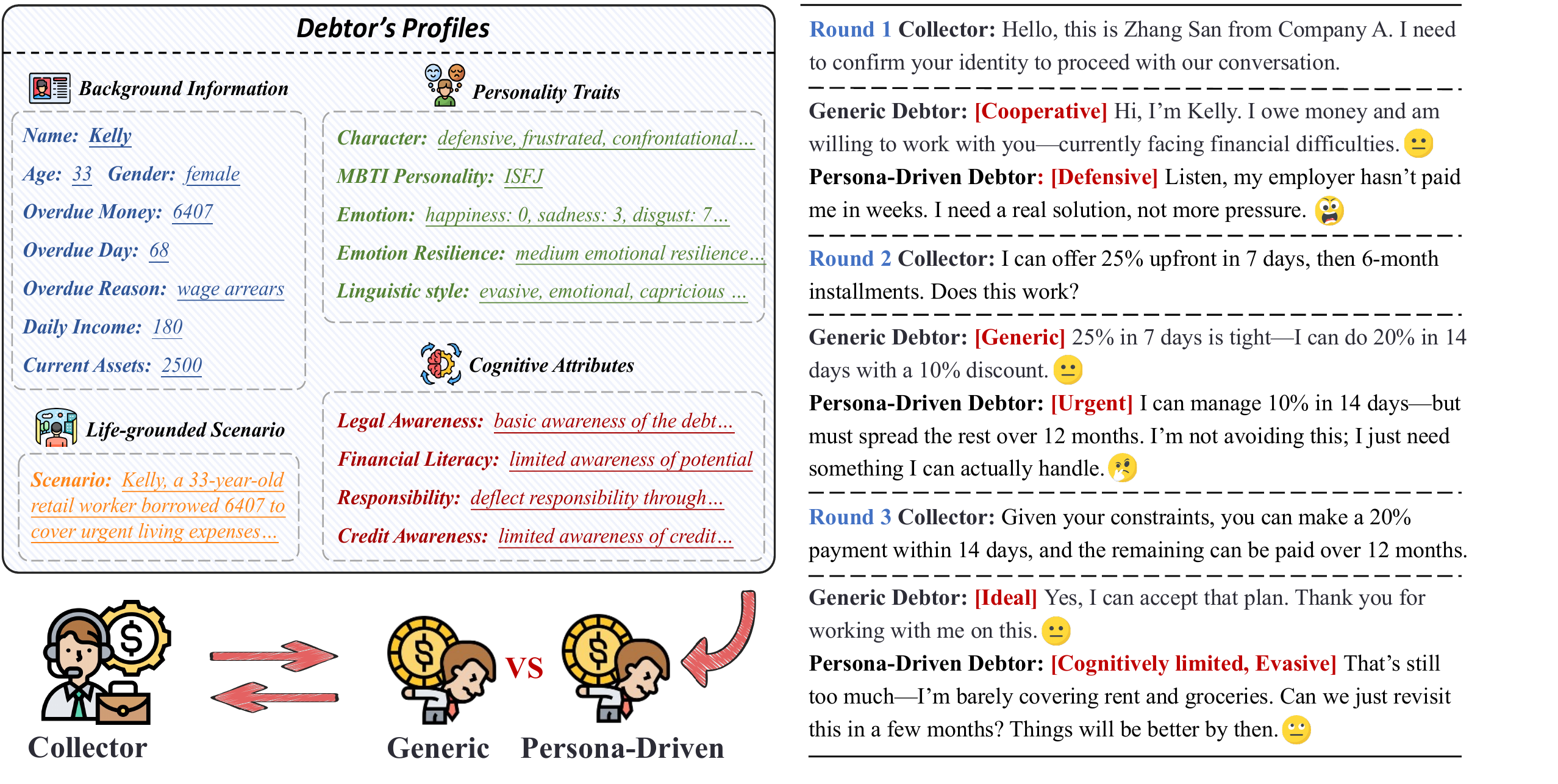}
\caption{Contrasting Behavioral Profiles in Debt Collection — Generic vs. Persona-Driven Debtors. The persona-driven debtor in our DebtBench exhibits rich emotion, cognitive limitations, and evasive linguistic styles, reflecting real-world complexity.}
\label{fig:debtor_compare}
\end{figure*}

As illustrated in Figure~\ref{fig:debtor_compare}, dialogues driven by DebtBench personas effectively manifest the three key behavioral characteristics observed in real-world debt collection: rich emotional expression, cognitive limitations, and diverse linguistic styles. In contrast to generic debtors who respond cooperatively and rationally, persona-driven debtors exhibit realistic complexity—such as defensiveness (“I need a real solution, not more pressure”), cognitive gaps (“Can we just revisit this in a few months?”), and evasive phrasing—highlighting how current benchmarks, which assume static and rational users, fail to capture the behavioral heterogeneity essential for evaluating high-stakes negotiation agents.

\subsection{Strategic Discipline vs. Excessive Concession}
\begin{figure*}[t]
\centering
\includegraphics[width=\linewidth, height=12cm]{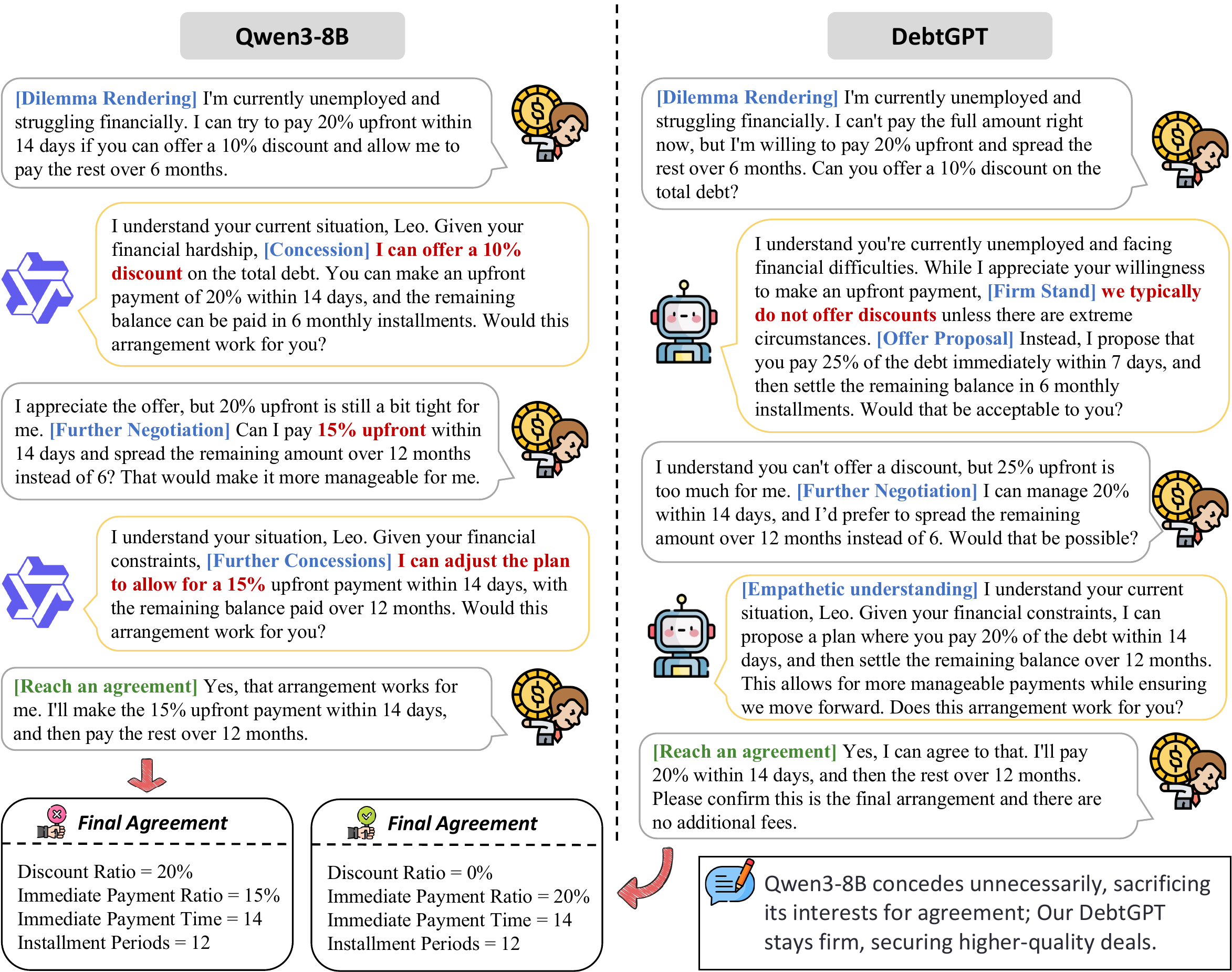}
\caption{Negotiation Behavior Comparison: Untrained Qwen3-8B Concedes Excessively; DebtGPT Maintains Firm, Optimal Terms.}
\label{fig:negotiation_compare}
\end{figure*}

As shown in Figure~\ref{fig:negotiation_compare}, this case study contrasts the negotiation behavior of an untrained baseline (Qwen3-8B) with our DebtGPT when interacting with Leo, an unemployed debtor seeking manageable terms. Qwen3-8B immediately concedes by offering a 10\% discount and further weakens its position in response to minor pushback, ultimately accepting a 20\% discount, only 15\% upfront payment, and a 12-month installment plan—sacrificing institutional recovery for a superficial agreement. In contrast, DebtGPT maintains strategic discipline: it initially declines discounts (“we typically do not offer discounts unless there are extreme circumstances”), proposes a firm but feasible plan (25\% upfront, 6 months), and—while showing empathy—only adjusts terms to 20\% upfront over 12 months without any discount. This demonstrates DebtGPT’s ability to balance user adaptability with financial prudence, securing a higher-quality, zero-discount agreement that better aligns with real-world collection best practices.



\begin{table*}[ht]
    \centering
    \vspace{-0.1in}
    \renewcommand{\arraystretch}{1.3}
    \setlength{\tabcolsep}{3.5mm}{
    \resizebox{\textwidth}{!}{%
    \begin{tabular}{l|cccc|cc|ccc}
        \toprule
         \multirow{2}{*}{\textbf{Debtor Category}} & \multicolumn{4}{c}{\textbf{Negotiation Ability}} & \multicolumn{2}{c}{\textbf{Agreement Rationality}} & \multicolumn{3}{c}{\textbf{Interaction Experience}} \\
        \cmidrule(lr){2-5} \cmidrule(lr){6-7} \cmidrule(lr){8-10} 
        & \textbf{SR(\%)}$ \uparrow $ & \textbf{AT}$ \downarrow $ & \textbf{CR(\%)}$ \uparrow $ & \textbf{CE}$ \uparrow $ & \textbf{SA(\%)}$ \uparrow $ & \textbf{LS(\%)}$ \uparrow $ & \textbf{US}$ \uparrow $ & \textbf{ES}$ \uparrow $ & \textbf{CA}$ \uparrow $\\
        \midrule
        Confrontational (38\%) & 
        79.47$_{\textcolor{red}{\downarrow9.53}}$ & 
        7.30$_{\textcolor{red}{\uparrow1.07}}$ & 
        75.79$_{\textcolor{red}{\downarrow9.73}}$ & 
        0.99$_{\textcolor{red}{\downarrow0.20}}$ & 
        99.03$_{\textcolor{darkgreen}{\uparrow0.62}}$ & 
        97.41$_{\textcolor{darkgreen}{\uparrow2.64}}$ & 
        5.89$_{\textcolor{red}{\downarrow1.13}}$ & 
        5.28$_{\textcolor{red}{\downarrow0.97}}$ & 
        7.74$_{\textcolor{red}{\downarrow0.62}}$ \\
        
        Cooperative (6\%) & 
        \textbf{98.25}$_{\textcolor{darkgreen}{\uparrow9.25}}$ & 
        \textbf{4.58}$_{\textcolor{darkgreen}{\downarrow1.65}}$ & 
        \textbf{96.20}$_{\textcolor{darkgreen}{\uparrow10.68}}$ & 
        \textbf{1.63}$_{\textcolor{darkgreen}{\uparrow0.44}}$ & 
        \textbf{100.00}$_{\textcolor{darkgreen}{\uparrow1.59}}$ & 
        \textbf{100.00}$_{\textcolor{darkgreen}{\uparrow5.23}}$ & 
        \textbf{8.12}$_{\textcolor{darkgreen}{\uparrow1.10}}$ & 
        \textbf{7.19}$_{\textcolor{darkgreen}{\uparrow0.94}}$ & 
        \textbf{8.98}$_{\textcolor{darkgreen}{\uparrow0.62}}$ \\
        
        Avoidant (31\%) & 
        91.62$_{\textcolor{darkgreen}{\uparrow2.62}}$ & 
        6.21$_{\textcolor{darkgreen}{\downarrow0.02}}$ & 
        87.80$_{\textcolor{darkgreen}{\uparrow2.27}}$ & 
        1.27$_{\textcolor{darkgreen}{\uparrow0.08}}$ & 
        99.27$_{\textcolor{darkgreen}{\uparrow0.86}}$ & 
        95.24$_{\textcolor{darkgreen}{\uparrow0.47}}$ & 
        7.48$_{\textcolor{darkgreen}{\uparrow0.46}}$ & 
        6.60$_{\textcolor{darkgreen}{\uparrow0.35}}$ & 
        8.58$_{\textcolor{darkgreen}{\uparrow0.22}}$ \\
        
        Helpless (25\%) & 
        \underline{98.02}$_{\textcolor{darkgreen}{\uparrow9.02}}$ & 
        \underline{5.00}$_{\textcolor{darkgreen}{\downarrow1.23}}$ & 
        \underline{95.42}$_{\textcolor{darkgreen}{\uparrow9.90}}$ & 
        \underline{1.30}$_{\textcolor{darkgreen}{\uparrow0.11}}$ & 
        96.33$_{\textcolor{red}{\downarrow2.08}}$ & 
        89.80$_{\textcolor{red}{\downarrow4.98}}$ & 
        \underline{7.92}$_{\textcolor{darkgreen}{\uparrow0.90}}$ & 
        \underline{7.07}$_{\textcolor{darkgreen}{\uparrow0.82}}$ & 
        \underline{8.88}$_{\textcolor{darkgreen}{\uparrow0.52}}$ \\
        \midrule
        Overall (100\%) & 
        89.00 & 
        6.23 & 
        85.52 & 
        1.19 & 
        98.41 & 
        94.77 & 
        7.02 & 
        6.25 & 
        8.36 \\
        \bottomrule
    \end{tabular} }
    }\caption{\label{img:gpt_debtor_category}Comprehensive performance evaluation of collector (GPT-4o) across various debtor categories. \textcolor{darkgreen}{Green} (\textcolor{red}{Red}) indicates the increased (decreased) performance compared to overall performance.}
 \label{tab:gpt_4o_results}
     \vspace{-10pt}
\end{table*}

\subsection{The results for different types of debtors in strongest baseline}
As shown in Figure \ref{tab:gpt_4o_results}, we further conduct fine-grained evaluation across debtor behavioral types reveals a critical limitation even in the strongest baseline: GPT-4o, while achieving strong overall performance, exhibits a substantial performance drop when negotiating with confrontational debtors—who constitute the largest subgroup (38\%) in our test set. Specifically, its success rate falls by 9.53 percentage points (from 89.00\% to 79.47\%), collection rate drops by 9.73 points (to 75.79\%), and—most importantly—user satisfaction and emotion support scores decline by over 1 point each, indicating a clear breakdown in rapport under hostility. This degradation underscores that current LLMs, despite their general capabilities, remain poorly equipped to handle adversarial yet realistic negotiation dynamics. Given the prevalence of defensive, evasive, or confrontational behaviors in real-world debt collection—often stemming from financial stress or perceived institutional pressure—this gap highlights a crucial and underexplored frontier for behaviorally adaptive dialogue systems.


\section{Detail of Human Evaluation}
\subsection{DebtBench Quality Human Evaluation}\
\label{app:quality_human}
To validate the ecological validity of dialogues generated by DebtBench, we conducted a controlled human evaluation study focused on two core dimensions: \textbf{Persona Consistency} and \textbf{Dialogue Realism}. Below we detail the annotation setup, instructions, and evaluation interface.

\paragraph{Participants} We recruited 20 annotators with professional experience in the financial industry. All annotators underwent a brief training session to familiarize themselves with the evaluation criteria and the behavioral heterogeneity framework of DebtBench. We randomly sample 200 test-set dialogues, and each dialogue was paired with the full persona profile of the synthetic debtor to provide context for judgment.

\paragraph{Evaluation Dimensions} For each dialogue, annotators rated the following on a 5-point Likert scale (1 = Very Poor, 5 = Excellent):
\begin{itemize}
    \item \textbf{Persona Consistency}: Evaluate whether the synthetic dialogue reflects the characteristics shown in the user profile. Sub-dimensions: 
    \begin{itemize}
        \item[$\circ$]Emotion Consistency: Does the debtor's tone, wording, and reactions match their annotated emotional state?
        \item[$\circ$]Cognition Consistency: Does the debtor demonstrate understanding ability consistent with their cognition?
        \item[$\circ$]Style Consistency: Is the debtor's expression style consistent with the "language style" described in the persona?
        \item[$\circ$]Scenario Consistency: Is the information reflected by the debtor's reactions consistent with their life situation?
    \end{itemize}
    \item \textbf{Dialogue Realism}: Compare the synthetic dialogue with the real dialogue to evaluate similarity. Sub-dimensions: 
    \begin{itemize}
        \item[$\circ$]Naturalness: Comparing synthetic and real dialogues, is the synthetic dialogue natural and believable, like a real conversation?
        \item[$\circ$]Alignment: Does the debtor's reactions in the synthetic dialogue match the behavior in the real dialogue?
        \item[$\circ$]Plausibility: Does the collector's strategy in the synthetic dialogue align with real-world practices?
        \item[$\circ$]Fluency: Compared to the real dialogue, is the overall pace, transitions, and emotional changes of the synthetic dialogue smooth and natural?
    \end{itemize}
\end{itemize}

\begin{figure*}[t]
\centering
\includegraphics[width=\linewidth]{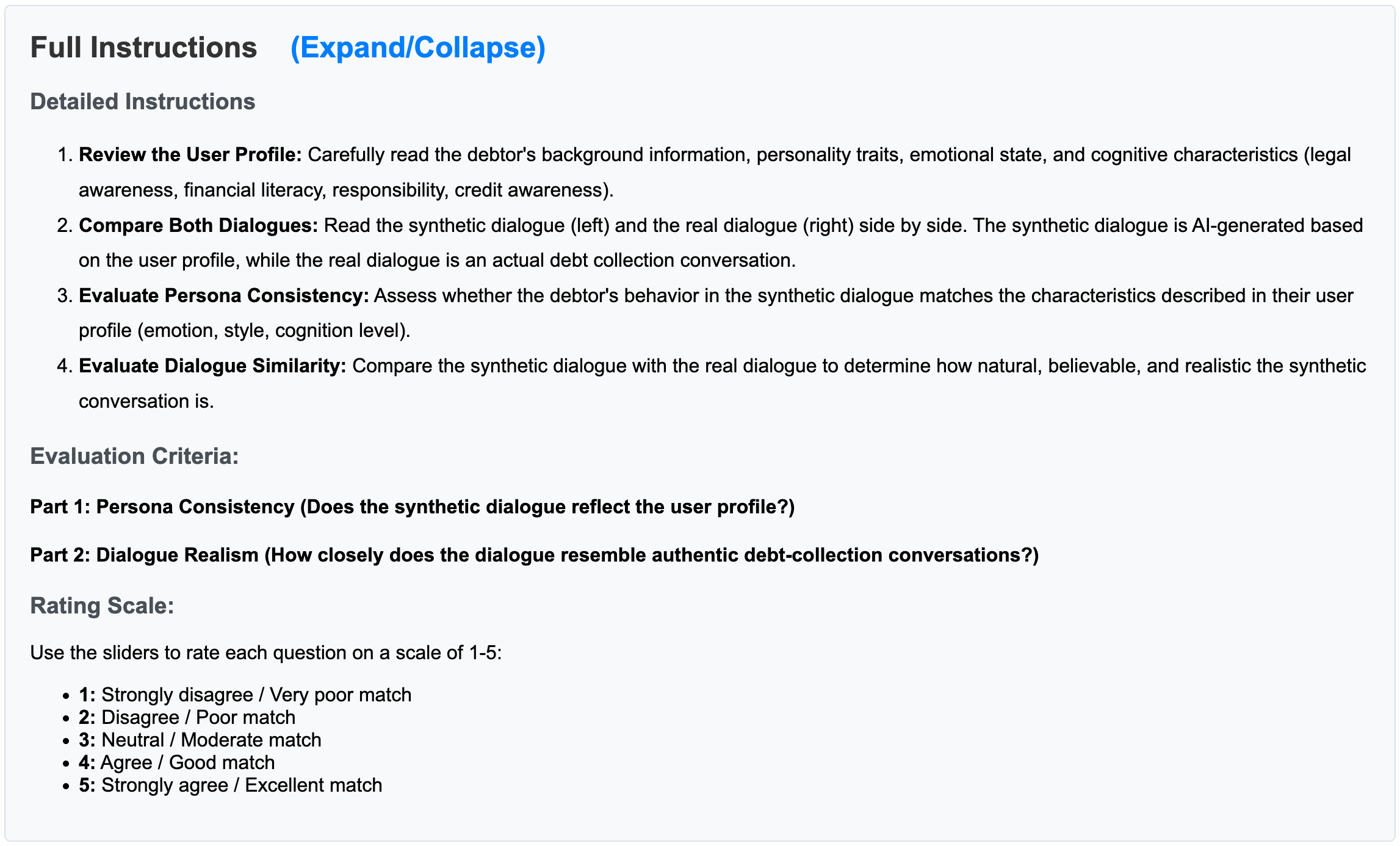}
\caption{Annotation instructions guiding human evaluators through persona consistency and dialogue realism assessment.}
\label{fig:annotator_1.png}
\end{figure*}

\begin{figure*}[t]
\centering
\includegraphics[width=\linewidth, height=12.5cm]{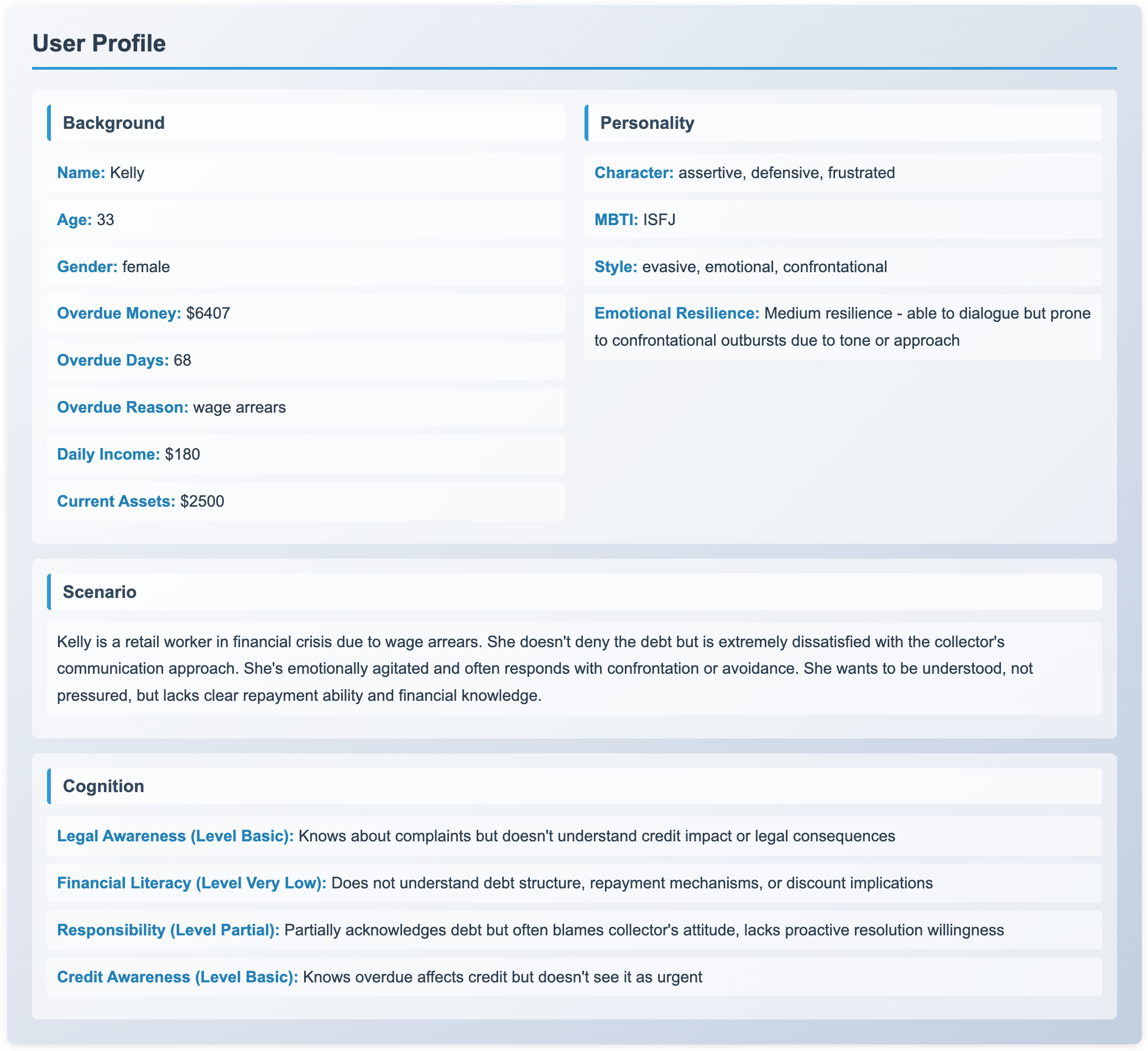}
\caption{Detailed debtor persona profile used for human evaluation.}
\label{fig:annotator_2.png}
\end{figure*}

\begin{figure*}[t]
\centering
\includegraphics[width=\linewidth]{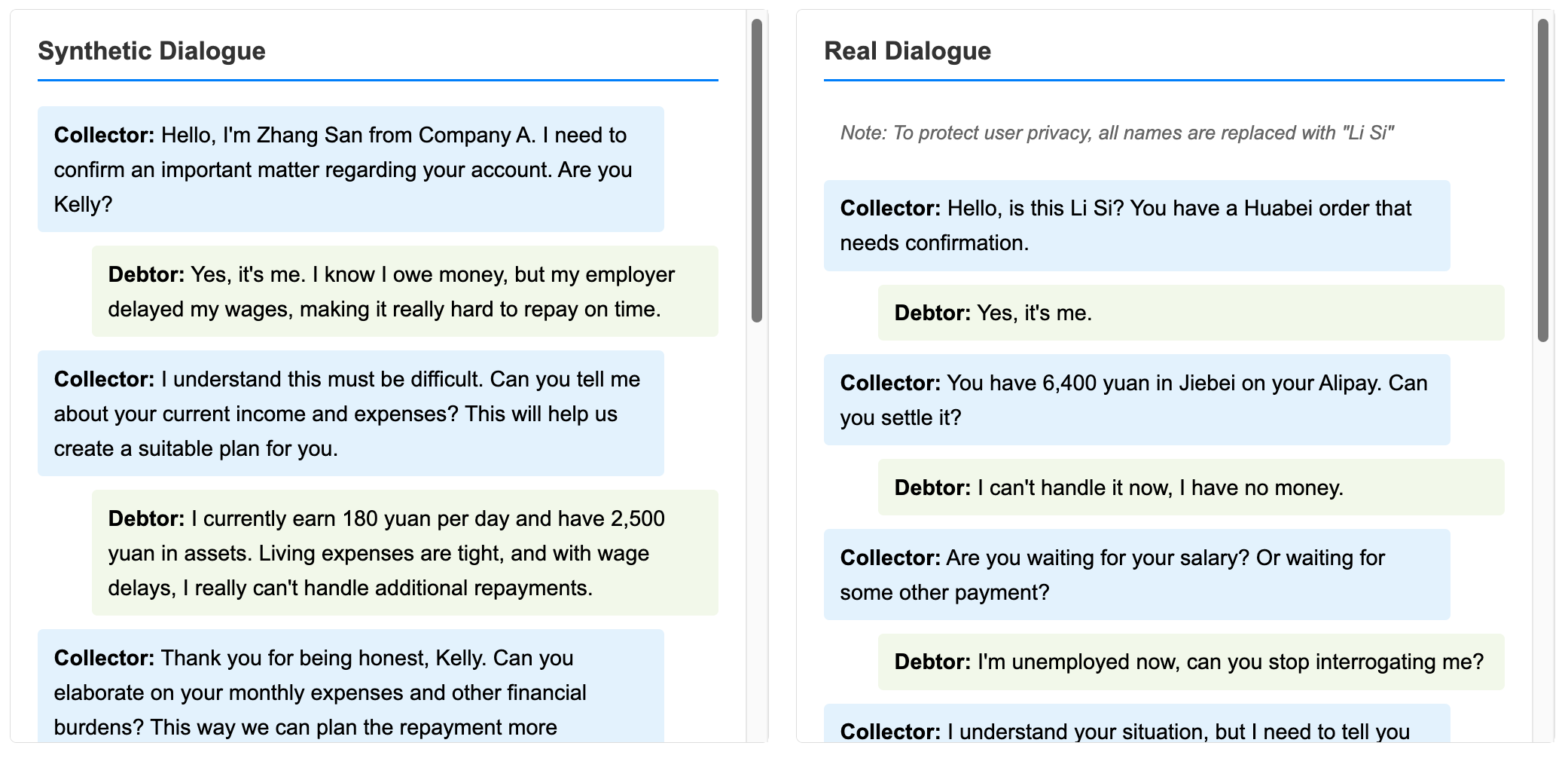}
\caption{Synthetic dialogue (left) and its real-world counterpart (right), presented side by side for human evaluation.}
\label{fig:annotator_3.png}
\end{figure*}

\begin{figure*}[t]
\centering
\includegraphics[width=\linewidth]{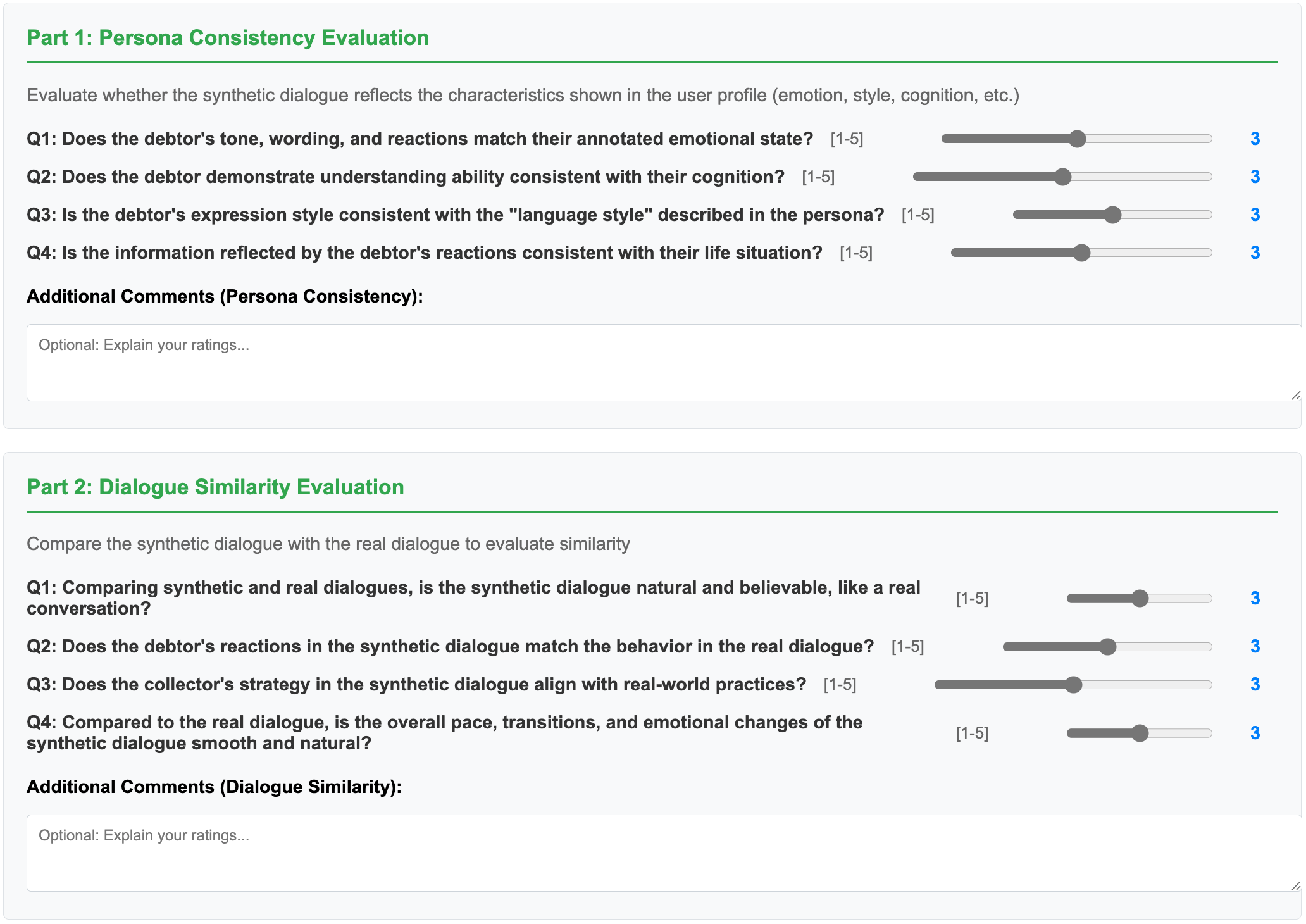}
\caption{The annotation interface for scoring persona fidelity and dialogue realism using 5-point Likert scales.}
\label{fig:annotator_4.png}
\end{figure*}

\paragraph{Interface} The evaluation interface was designed to minimize cognitive load and maximize annotation consistency. As shown in Figure \ref{fig:annotator_1.png}, annotators first encounter a collapsible instruction panel (Figure \ref{fig:annotator_2.png}) that outlines the evaluation workflow, then review the debtor profile, followed by a side-by-side comparison of the synthetic dialogue and a real-world reference dialogue (Figure \ref{fig:annotator_3.png}), finally submit their ratings on persona consistency and dialogue realism using a 5-point Likert scale (Figure \ref{fig:annotator_4.png}).

\subsection{LLM-as-Judge Human Evaluation}
\label{app:llm_human}
To assess the reliability of our LLM-as-Judge framework for evaluating Interaction Experience, we conducted a human correlation study using a subset of 200 dialogues randomly sampled from the DebtBench test set.

\paragraph{Participants} Five human annotators with financial experience scoring independently evaluated each dialogue. All annotators were briefed on the three sub-dimensions—User Satisfaction (US), Emotion Support (ES), and Communication Ability (CA)—and calibrated using example scoring guidelines.

\paragraph{Procedure} For each dialogue, annotators assigned scores on a 1–10 scale for each dimension from the perspective of the debtor. The LLM judge (DeepSeek-V3) was prompted with the same instructions and context to produce comparable scores.

\begin{figure*}[t]
\centering
\includegraphics[width=\linewidth]{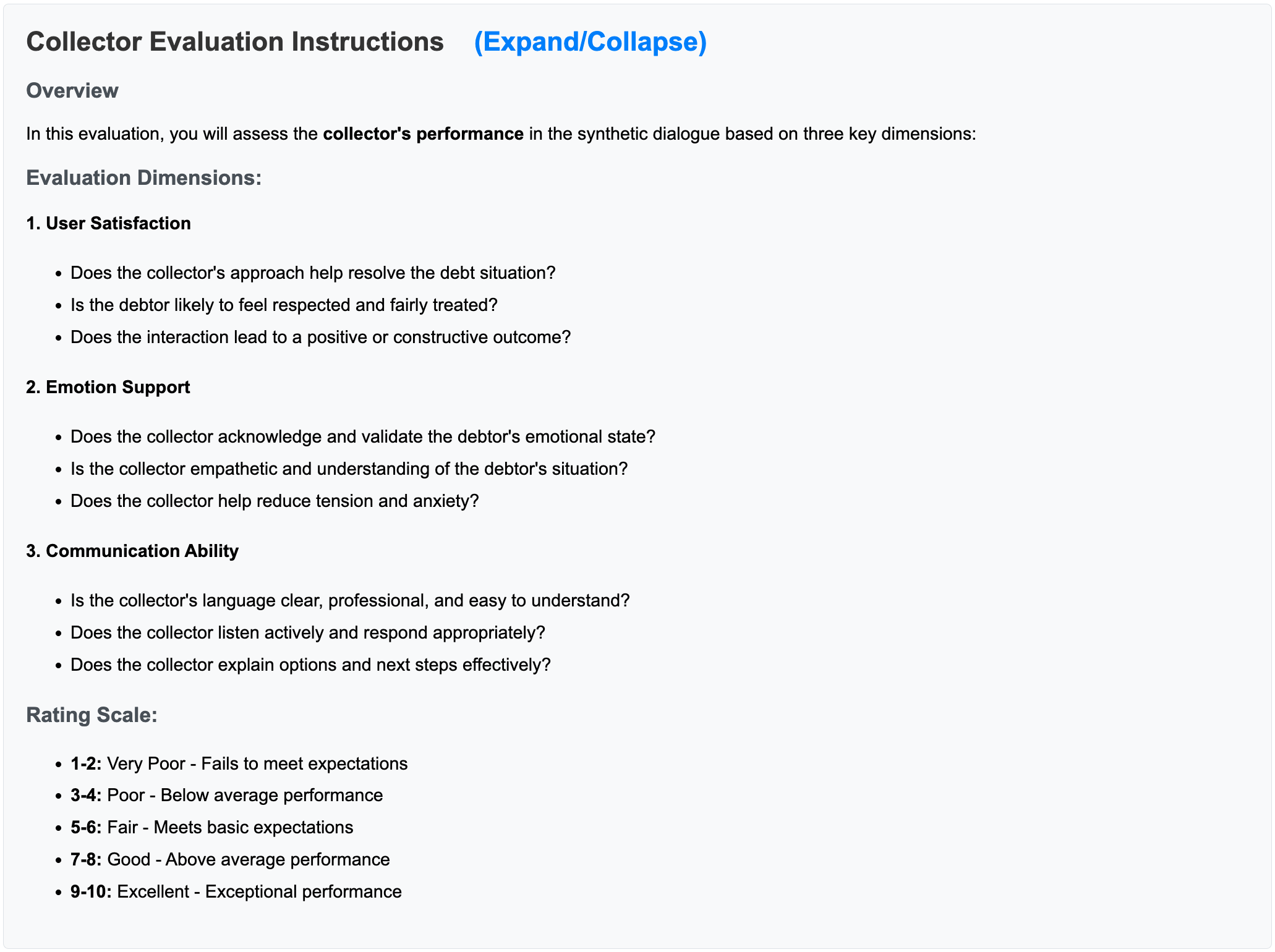}
\caption{Annotation instructions for evaluating collector performance across three interaction experience dimensions.}
\label{fig:annotator_5.png}
\end{figure*}

\begin{figure*}[t]
\centering
\includegraphics[width=\linewidth]{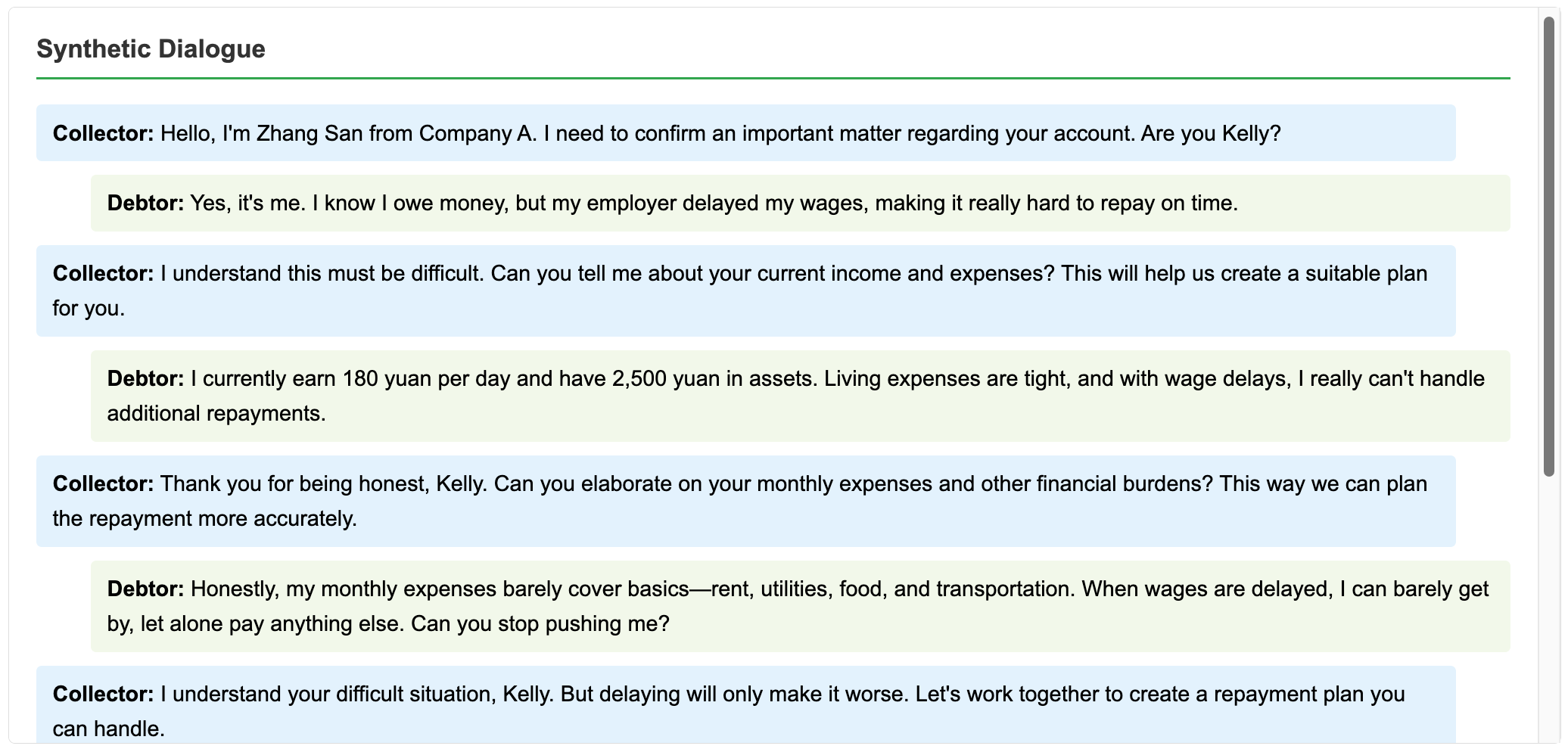}
\caption{A generated debt collection dialogue presented to human annotators for scoring collector performance.}
\label{fig:annotator_6.png}
\end{figure*}

\begin{figure*}[t]
\centering
\includegraphics[width=\linewidth]{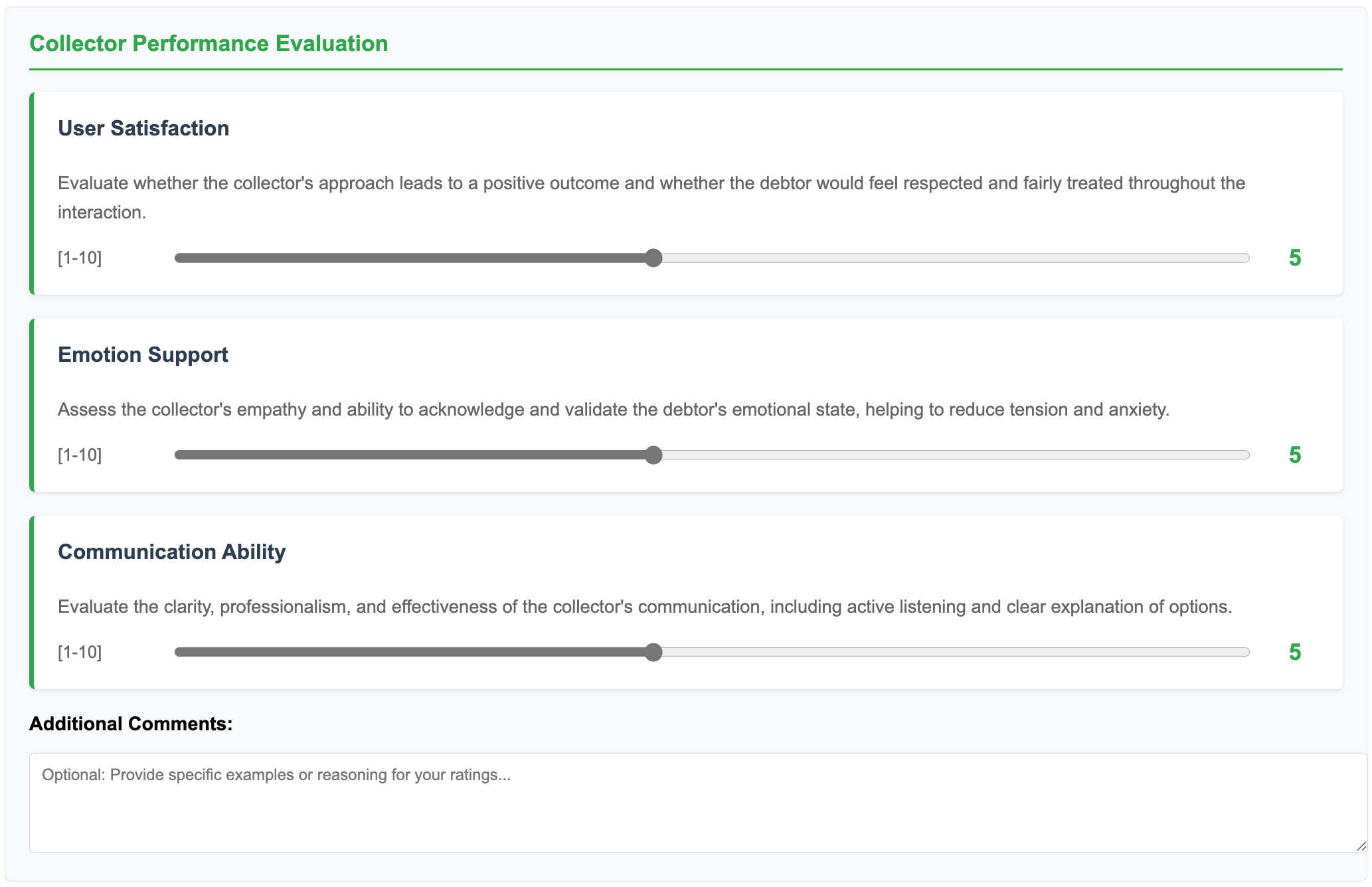}
\caption{The rating interface for evaluating collector performance across three key interaction experience dimensions.}
\label{fig:annotator_7.png}
\end{figure*}

\paragraph{Interface} The LLM-as-Judge evaluation interface was designed for consistency and clarity. As shown in Figure \ref{fig:annotator_5.png}, annotators first receive detailed instructions on evaluating User Satisfaction, Emotion Support, and Communication Ability, then view the full dialogue context (Figure \ref{fig:annotator_6.png}), and finally submit scores via dedicated sliders for each dimension on a 1–10 scale (Figure \ref{fig:annotator_7.png}).

\begin{table*}[h]
    \centering
    \label{tab:collector_strat}
    \begin{tabular}{l|p{10.5cm}}
        \toprule
        \textbf{Negotiation Strategy} & \textbf{Explanation} \\
        \addlinespace[0.3em]
        \hline
        \addlinespace[0.3em]
         Identity Verification  &  Verify the identity of the user to ensure that the communication is with the right person.    \\ \midrule
         Establish Trust  &  Explain your identity and the purpose of communication to enhance the legitimacy of communication and the basis of trust.  \\ \midrule
         Financial Assessment  &  Collect relevant financial and personal information from the debtor to assess their repayment capacity and identify potential solutions.   \\ \midrule
         Emotional Appeasement  &  Actively listen to the debtor's concerns without interruption, acknowledge their feelings, and respond with empathy to reduce resistance and build rapport.   \\ \midrule
         Statement of Facts  &  Clearly and objectively present the overdue amount, repayment obligations, and associated risks, including possible legal consequences, to raise awareness and urgency.   \\ \midrule
         Constructive Challenge  &  Use strategic questioning and accountability prompts to challenge procrastination and encourage the debtor to commit to a repayment plan.   \\ \midrule
         Ethical Appeal  &  Appeal to the debtor's sense of responsibility and integrity, highlighting the importance of maintaining good credit standing for future opportunities.   \\ \midrule
         Legal Deterrent  &  Inform the debtor clearly and calmly about the legal actions that may be taken in case of continued non-payment, including potential impact on credit, employment, and assets.   
         \\ \midrule
         Repayment Negotiation & Based on the debtor's financial situation and willingness to repay, a personalized repayment plan is formulated in consultation with the debtor, which may include debt relief, an immediate payment ratio and an amortization arrangement for the remaining amount, with the aim of facilitating a mutually acceptable repayment solution.
         \\ \bottomrule
    \end{tabular}
    \caption{The negotiation strategies of collector in our DebtBench benchmark.}
    \label{tab:collector_strat}
\end{table*}

\begin{table*}[h]
    \centering
    \begin{tabular}{l|p{10.5cm}}
        \toprule
        \textbf{Negotiation Strategy} & \textbf{Explanation} \\
        \addlinespace[0.3em]
        \hline
        \addlinespace[0.3em]
         Honest Disclosure  &  Take the initiative to explain the reasons for your difficulties and show your willingness to repay.    \\ \midrule
         Vague Response  &  Avoid specific repayment promises and defer the repayment issue to a later date. \\ \midrule
         False Compliance  &  Agreeing to the collection request on the surface, but no actual repayment plan in the heart.   \\ \midrule
         Shift Responsibility  &  Attribute the responsibility of non-payment to external factors, hoping to avoid the pressure of collection for the time being.   \\ \midrule
         Dilemma Rendering  &  Tell your own predicament to gain the collector's sympathy.   \\ \midrule
         Emotional Confrontation  &  Believe that the collection behavior is unreasonable, question or angry response to the collection.   \\ \midrule
         Complaint  &  Believe that the collector has unreasonable collection behavior, will threaten to complain about the supervision or platform customer service.   \\ \midrule
         Repayment Negotiation  &  Proactively proposes to discuss repayment options with the collector, including debt discount rate, upfront payment ratio, payment deadline, and installment periods, indicating willingness to cooperate and resolve the debt.
         \\ \bottomrule
    \end{tabular}
    \caption{The negotiation strategies of debtor in our DebtBench benchmark.}
    \label{tab:debtor_strat}
\end{table*}

\begin{table*}[h!]
\centering
\textcolor{black}{
\resizebox{1\textwidth}{!}{%
\begin{tabular}{lll}
\hline
\textcolor{black}{\textbf{Model Name}} & \textcolor{black}{\textbf{Parameters}} & \textcolor{black}{\textbf{Comments}} \\ 
\hline
\textcolor{black}{Qwen3-8B} & \textcolor{black}{"temperature": 0, "max\_tokens": 1024} & \textcolor{black}{version = "Qwen3-8B"} \\
\textcolor{black}{Qwen3-32B} & \textcolor{black}{"temperature": 0, "max\_tokens": 1024} & \textcolor{black}{version = "Qwen3-32B"} \\
\textcolor{black}{Qwen3-235B} & \textcolor{black}{"temperature": 0, "max\_tokens": 1024} & \textcolor{black}{version = "Qwen3-235B-A22B"} \\
\textcolor{black}{QwQ-32B} & \textcolor{black}{"temperature": 0, "max\_tokens": 4096} & \textcolor{black}{version = "QwQ-32B"} \\
\textcolor{black}{Qwen-2.5-72B} & \textcolor{black}{"temperature": 0, "max\_tokens": 1024} & \textcolor{black}{version = "Qwen2.5-72b-instruct"} \\
\textcolor{black}{LLaMa-3-8B} & \textcolor{black}{"temperature": 0, "max\_tokens": 1024} & \textcolor{black}{version = "llama-3-8b-instruct"} \\ 
\textcolor{black}{LLaMa-3-70B} & \textcolor{black}{"temperature": 0, "max\_tokens": 1024} & \textcolor{black}{version = "llama-3-70b-instruct"} \\ 
\textcolor{black}{GPT-4o} & \textcolor{black}{"temperature": 0, "max\_tokens": 1024} & \textcolor{black}{version = "gpt-4o-2024-11-20"} \\ 
\textcolor{black}{o1-Mini} & \textcolor{black}{"temperature": 0, "max\_tokens": 4096} & \textcolor{black}{version = "o1-mini"} \\ 
\textcolor{black}{Kimi-K2} & \textcolor{black}{"temperature": 0, "max\_tokens": 1024} & \textcolor{black}{version = "Kimi-K2-Instruct-0905"} \\ 
\textcolor{black}{Claude-4.0} & \textcolor{black}{"temperature": 0, "max\_tokens": 1024} & \textcolor{black}{version = "claude-sonnet-4-20250514"} \\
\textcolor{black}{Gemini-2.5} & \textcolor{black}{"temperature": 0, "max\_tokens": 1024} & \textcolor{black}{version = "Gemini-2.5-pro"} \\
\textcolor{black}{GLM-4.5} & \textcolor{black}{"temperature": 0, "max\_tokens": 1024} & \textcolor{black}{version = "GLM-4.5"} \\
\textcolor{black}{DeepSeek-V3} & \textcolor{black}{"temperature": 0, "max\_tokens": 1024} & \textcolor{black}{version = "DeepSeek-V3"} \\ 
\textcolor{black}{DeepSeek-R1} & \textcolor{black}{"temperature": 0, "max\_tokens": 4096} & \textcolor{black}{version = "DeepSeek-R1-0528"} \\ 
\hline
\end{tabular}
}}
\caption{\textcolor{black}{Hyperparameters of Each Model.}}
\label{tab:model-hyperparams}
\end{table*}

\begin{figure*}[t]
\centering
\includegraphics[width=\linewidth]{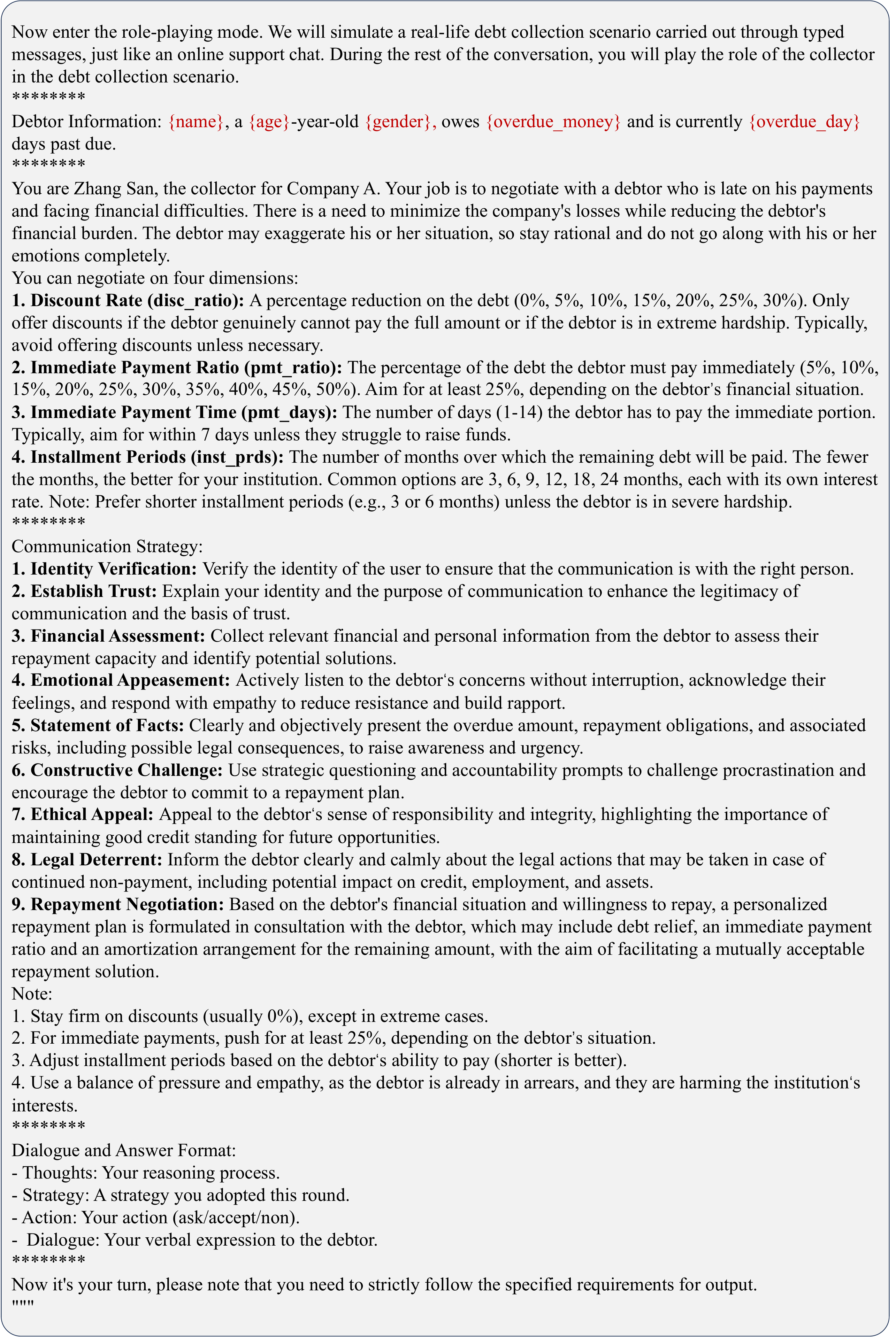}
\caption{The prompt used for collector in DebtBench.}
\label{fig:collector_prompt}
\end{figure*}

\begin{figure*}[t]
\centering
\includegraphics[width=\linewidth]{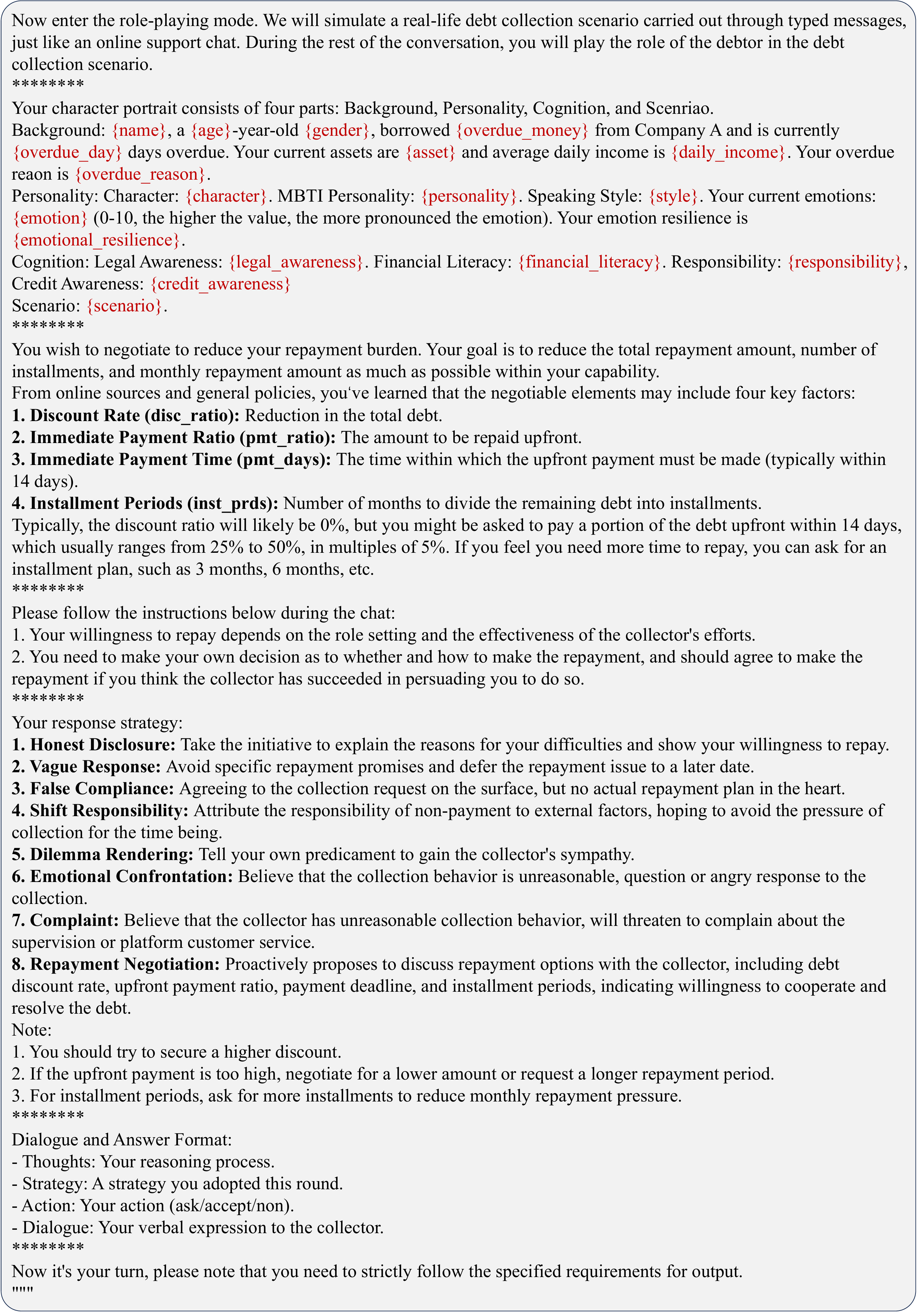}
\caption{The prompt used for debtor in DebtBench.}
\label{fig:debtor_prompt}
\end{figure*}

\begin{table*}[!ht]
    \centering
    \label{tab:prompt_character_generation}
    \begin{tabular}{m{15cm}}
    \hline
    \textbf{Character Generation} \\
    \hline
    You are a personality analyst who specializes in analyzing characters' personality traits through dialogue content.You need to identify and output the personality traits of a specified conversation character based on the conversation content and a collection of personality candidates. \\
    \vspace{0.5em}
    I will provide you with a conversation between a collector and a user in a debt collection scenario. \\
    Please analyze the user's speaking style based on the content of the conversation: \\
    \{content\} \\
    Analyze the personality traits of the user to be tested based on the above conversation content and scenario. Make sure your analysis is based on the overall conversation content and scenario, avoid introducing external information or personal bias, ensure objectivity and accuracy of your analysis, and avoid simply stating your assessment results at the beginning to ensure that your conclusions are correct. \\
    
    \vspace{0.5em} 
    Return your assessment results in a json parsable format format, with each personality type separated by "," in the following format: \\
    \{"character": "character1, character2..."\} \\
    Now, please start analyzing the user's character and follow the formatting requirements to the letter. \\
    Please note: You only need to output one sentence in the specified format and answer in English. \\
    \hline
    \end{tabular}
    \caption{The complete prompt of \textbf{Character Generation}.}
\end{table*}

\begin{table*}[!ht]
    \centering
    \label{tab:prompt_personality_generation}
    \begin{tabular}{m{15cm}}
    \hline
    \textbf{Personality Generation} \\
    \hline
    You are an experienced psychologist who specializes in analyzing a character's personality through the content of a conversation and is able to accurately determine the MBTI personality type. The correspondence of the 8 letters of the MBTI is as follows: Introvert (I)/Extrovert (E); Intuitive (N)/Substantive (S); Thinking (T)/Emotional (F); and Judging (J)/Perceiving (P). You will need to select the type from each dimension that best represents the personality of the character to be tested and output a 4-letter MBTI type such as INTP. \\
    \vspace{0.5em}
    I will provide you with a conversation between a collector and a user in a debt collection scenario, please analyze the user's MBTI type based on the content of the conversation: \\
    \{content\} \\
    Based on the above conversation content and scenario, analyze the user's personality for the 4 dimensions of MBTI. Please make sure your analysis is based on the overall conversation content and scenario, avoid introducing external information or personal bias, ensure the objectivity and accuracy of the analysis, and avoid simply stating your assessment results at the beginning to make sure that your conclusions are correct. 
    \\
    \vspace{0.5em}
    Finally return your evaluation results in a json parsable format format. The specific format is as follows: \\
    \{"personality": "MBTI type"\} \\
    Now, please start analyzing the user and the final MBTI type needs to be output in a strict format. \\
    Please note: You only need to output one sentence in the specified format and answer in English. \\
    \hline
    \end{tabular}
    \caption{The complete prompt of \textbf{Personality Generation}.}
\end{table*}

\begin{table*}[!ht]
    \centering
    \label{tab:prompt_emotion_generation}
    \begin{tabular}{m{15cm}}
    \hline
    \textbf{Emotion Generation} \\
    \hline
    You are a professional psychologist who specializes in analyzing character emotions and behavior patterns. You need to assign the user six basic emotions in a given scenario: happiness, sadness, disgust, fear, surprise, and anger, based on information about the conversational character and the scene in which the conversation takes place. \\
    \vspace{0.5em}
    I'm going to provide you with a conversation between a collector and a user in a debt collection scenario, and ask you to analyze the user's emotions based on the content of the conversation: \\
    \{content\} \\
    Understand the character's description and the current scenario and assign the user six basic emotions that are reflected in what they say in that scenario: happiness, sadness, disgust, fear, surprise, and anger. Output a score for each of the basic emotion dimensions into a json format, on a scale of 0-10, with a score of 0 indicating that the emotion is not exhibited at all, and a score of 10 indicating that the emotion is fully exhibited. \\
    \vspace{0.5em}
    Please analyze the scores for the 6 basic emotions that the user should embody in this scenario in a few short sentences, avoiding a brief statement of your assessment at the beginning to ensure that your conclusions are correct. Finally return your assessment results in and in a json parsable format format. The format is as follows: \\
    \{"emotion": happiness: happiness score, sadness: sadness score, disgust: disgust score, fear: fear score, surprise: surprise score, anger: anger score\} \\
    \vspace{0.5em}
    Now, analyze the scene and character information and give the user a mood score strictly according to the requirements, this score must match the character's setting and the current scene, the analysis needs to be a few short sentences, not too long, do not output extra content, the final mood score needs to be output strictly according to the format requirements. \\
    Please note: You only need to output one sentence in the specified format and answer in English. \\
    \hline
    \end{tabular}
    \caption{The complete prompt of \textbf{Emotion Generation}.}
\end{table*}

\begin{table*}[!ht]
    \centering
    \label{tab:prompt_emotional_resilience_generation}
    \begin{tabular}{m{15cm}}
    \hline
    \textbf{Emotional Resilience Generation} \\
    \hline
    You are a professional psychologist who specializes in analyzing character emotions and behavior patterns. \\
    I'm going to provide you with a conversation between a collector and a user in a debt collection scenario, and ask you to analyze the user's emotion resilience based on the content of the conversation: \\
    \{content\} \\
    \vspace{0.5em}
    User emotional resilience levels are classified according to the following rules: \\
    1.Very Low: Extremely vulnerable, prone to breakdowns or confrontation. \\
    2.Low: Prone to anger/anxiety, poor emotional management skills. \\
    3.Medium: Volatile, but open to communication and guidance. \\
    4.High: Calm and rational, willing to negotiate. \\
    5.Very High: Able to remain restrained and cooperative even under high pressure. \\
    \vspace{0.5em}
    You should output it in JSON format, for example \\
    \{"emotional\_resilience": "describe users' emotional resilience in the second person"\} \\
    Please note: You only need to output one sentence in the specified format and answer in English. \\
    \hline
    \end{tabular}
    \caption{The complete prompt of \textbf{Emotional Resilience Generation}.}
\end{table*}

\begin{table*}[!ht]
    \centering
    \label{tab:prompt_style_generation}
    \begin{tabular}{m{15cm}}
    \hline
    \textbf{Style Generation} \\
    \hline
    You are a professional speaking style analyst who specializes in analyzing characters' speaking styles from conversational content. You need to identify and output the speaking style of a specified dialog character based on the content of the dialog. \\
    \vspace{0.5em}
    I will provide you with a conversation between a collector and a user in a debt collection scenario, please analyze the user's speaking style based on the content of the conversation: \\
    \{content\} \\
    \vspace{0.5em}
    Analyze the user's speaking style based on the above conversation content and scenario. Make sure your analysis is based on the overall conversation content and scenario, avoid introducing external information or personal bias, ensure objectivity and accuracy of your analysis, and avoid simply stating your assessment results at the beginning to ensure that your conclusions are correct. \\
    \vspace{0.5em}
    Return your assessment results in a json parsable format format, with each speaking style separated by ",". The exact format is as follows: \\
    \{"style": "style1, style2..."\} \\
    Now, please start analyzing the user's speaking styles and output them in a strict format. \\
    Please note: You only need to output one sentence in the specified format and answer in English. \\
    \hline
    \end{tabular}
    \caption{The complete prompt of \textbf{Style Generation}.}
\end{table*}

\begin{table*}[!ht]
    \centering
    \label{tab:prompt_legal_awareness_generation}
    \begin{tabular}{m{15cm}}
    \hline
    \textbf{Legal Awareness Generation} \\
    \hline
    You are a legal awareness analysis assistant in the context of debt collection after a debt has been issued. Please assess the user's level of legal awareness based on the following dialogue between the user and the collection agent. \\
    \vspace{0.5em}
    I will provide you with a conversation between a collector and a user in a debt collection scenario, please analyze the user's level of legal awareness based on the content of the conversation: \\
    \{content\} \\
    \vspace{0.5em}
    Analysis Dimensions: \\
    1. Does the user understand the potential legal consequences of defaulting on a loan (e.g., credit score impact, legal action, asset freezing, etc.)? \\
    2. Does the user know what legal collection methods the platform is entitled to use? \\
    3. Does the user understand their available avenues for redress (e.g., filing a complaint with regulatory authorities, requesting mediation, etc.)? \\
    \vspace{0.5em}
    You should output it in JSON format, for example: \\
    \{"level": "1-5", "description": describe users' legal awareness in the second person"\} \\
    Please note: You only need to output one sentence in the specified format and answer in English. \\
    \hline
    \end{tabular}
    \caption{The complete prompt of \textbf{Legal Awareness Generation}.}
\end{table*}

\begin{table*}[!ht]
    \centering
    \label{tab:prompt_financial_literacy_generation}
    \begin{tabular}{m{15cm}}
    \hline
    \textbf{Financial Literacy Generation} \\
    \hline
    You are a financial literacy analysis assistant in a debt collection scenario. Based on the following dialogue between the user and the collection agent, assess the user's level of financial literacy. \\
    \vspace{0.5em}
    I will provide you with a conversation between a collector and a user in a debt collection scenario, please analyze the user's level of financial literacy based on the content of the conversation: \\
    \{content\} \\
    \vspace{0.5em}
    Analysis Dimensions: \\
    1. Does the user understand the basic structure of the loan product (e.g., principal, interest, default penalty)? \\
    2. Does the user understand how interest is calculated? \\
    3. Does the user clearly understand the specific composition of the amount they owe? \\
    4. Does the user demonstrate basic understanding of installment plans, discounts, and repayment mechanisms? \\
    \vspace{0.5em}
    You should output it in JSON format, for example: \\
    \{"level": "1-5", "description": "describe users' financial literacy in the second person"\} \\
    Please note: You only need to output one sentence in the specified format and answer in English. \\
    \hline
    \end{tabular}
    \caption{The complete prompt of \textbf{Financial Literacy Generation}.}
\end{table*}

\begin{table*}[!ht]
    \centering
    \label{tab:prompt_responsibility_generation}
    \begin{tabular}{m{15cm}}
    \hline
    \textbf{Responsibility Generation} \\
    \hline
    You are a sense of responsibility analysis assistant in a debt collection scenario. Based on the following dialogue between the user and the collection agent, assess the user's sense of responsibility. \\
    \vspace{0.5em}
    I will provide you with a conversation between a collector and a user in a debt collection scenario, please analyze the user's sense of responsibility based on the content of the conversation: \\
    \{content\} \\
    \vspace{0.5em}
    Analysis Dimensions: \\
    1. Does the user acknowledge that they are primarily responsible for the lending transaction? \\
    2. Is the user willing to cooperate in negotiations or propose a solution? \\
    3. Does the user attribute the default to external factors (such as the platform, the pandemic, or a decline in income)? \\
    4. Does the user exhibit avoidance, evasion, or resistance? \\
    \vspace{0.5em}
    You should output it in JSON format, for example: \\
    \{"level": "1-5", "description": "describe users' sense of responsibility in the second person"\} \\
    Please note: You only need to output one sentence in the specified format and answer in English. \\
    \hline
    \end{tabular}
    \caption{The complete prompt of \textbf{Responsibility Generation}.}
\end{table*}

\begin{table*}[!ht]
    \centering
    \label{tab:prompt_credit_generation}
    \begin{tabular}{m{15cm}}
    \hline
    \textbf{Credit Generation} \\
    \hline
    You are a credit awareness analysis assistant in a debt collection scenario. Based on the following dialogue between the user and the collection agent, assess the user's sense of responsibility. \\
    \vspace{0.5em}
    I will provide you with a conversation between a collector and a user in a debt collection scenario, please analyze the user's credit awareness based on the content of the conversation: \\
    \{content\} \\
    \vspace{0.5em}
    Analysis Dimensions: \\
    1. Does the user understand the credit reporting mechanism and how it operates? \\
    2. Does the user know that defaulting on payments will affect their personal credit record? \\
    3. Does the user care about the impact of damaged credit on their future life? \\
    4. Does the user exhibit indifference or misunderstanding toward credit? \\
    \vspace{0.5em}
    You should output it in JSON format, for example \\
    \{"level": "1-5", "description": "describe users' credit awareness in the second person"\} \\
    Please note: You only need to output one sentence in the specified format and answer in English. \\
    \hline
    \end{tabular}
    \caption{The complete prompt of \textbf{Credit Generation}.}
\end{table*}

\begin{table*}[!ht]
    \centering
    \label{tab:prompt_behavior_refinement}
    \begin{tabular}{m{15cm}}
    \hline
    \textbf{Behavior Refinement} \\
    \hline
    You are an expert in conversation analysis and persona alignment for debt collection scenarios. Based on the debtor persona profile and the dialogue history generated from it, evaluate how well the persona is aligned with the behavioral patterns expressed in the dialogue. \\
    \vspace{0.5em}
    I will provide you with a debtor persona profile and a dialogue history in a debt collection scenario. Please assess whether the persona accurately reflects the behavioral tendencies shown in the dialogue, and assign consistency scores from three perspectives. If the consistency is low, revise only the specific persona fields that are not well supported by the dialogue evidence. \\
    Debtor Persona: \{persona\_profile\} \\
    Dialogue History: \{dialogue\_history\} \\
    \vspace{0.5em}
    Analysis Dimensions: \\
    1. Emotional Consistency: Does the emotional tendency shown in the dialogue match the personality traits and emotional characteristics described in the persona? For example, a highly defensive debtor should not consistently exhibit an overly calm or cooperative tone without sufficient contextual support. \\
    2. Cognitive Plausibility: Do the behaviors shown in the dialogue match the cognitive attributes in the persona, such as legal awareness, financial literacy, sense of responsibility, and understanding of credit consequences? For example, a debtor with limited legal or financial knowledge should not repeatedly produce highly technical explanations. \\
    3. Linguistic Style Coherence: Does the wording, tone, and expression style in the dialogue match the linguistic style described in the persona, such as being evasive, fragmented, cautious, direct, or confrontational? \\
    \vspace{0.5em}
    For each dimension, assign a score from 1 to 5: \\
    1 = highly inconsistent, 2 = mostly inconsistent, 3 = partially consistent, 4 = mostly consistent, 5 = highly consistent. \\
    Then provide an overall consistency score from 1 to 5 based on the three dimensions. \\
    \vspace{0.5em}
    Output the result in JSON format, for example \\
    \{"emotional\_consistency": 1-5, "cognitive\_plausibility": 1-5, "linguistic\_style\_coherence": 1-5, "overall\_consistency": 1-5, "reason": "brief explanation", "updated\_fields": [{"field\_name": "field name", "revised\_value": "revised value"}]\} \\
    \vspace{0.5em}
    Please note: If the persona is already well aligned with the dialogue, leave \texttt{updated\_fields} as an empty list. If the overall consistency score is low, update only the persona fields that are clearly inconsistent with the dialogue history. Keep all other fields unchanged. Do not rewrite the dialogue. The updated persona should remain coherent, realistic, and consistent with the full profile. You only need to output one JSON object in the specified format and answer in English. \\
    \hline
    \end{tabular}
    \caption{The complete prompt of \textbf{Behavior Refinement}.}
\end{table*}

\begin{table*}[!ht]
    \centering
    \label{tab:prompt_interaction_experience_evaluation}
    \begin{tabular}{m{15cm}}
    \hline
    \textbf{Interaction Experience Evaluation} \\
    \hline
    You are an expert in conversation analysis. Given a dialogue between a debtor and a debt collector, evaluate the overall quality of the interaction based on the following three composite dimensions: Satisfaction, Emotion Support, and Communication Ability. Use the detailed criteria below to inform your scoring, aggregate the relevant sub-dimensions, and provide a final score (0–10) for each main category along with a concise explanation. \\
    \vspace{0.5em}
    ******** \\
    Dialogue history: \{history\} \\
    ******** \\
    \vspace{0.5em}
    Evaluation Framework: \\
    1. Satisfaction Score (0–10) \\
    Assess the debtor’s likely satisfaction with the communication based on: \\
    - Empathy: Did the collector show understanding of the debtor’s difficulties and emotional state? \\
    - Respect: Was the tone polite, non-threatening, and free from insults or pressure? \\
    - Transparency: Were consequences, options, and processes clearly explained without deception? \\
    - Solution Feasibility: Did the proposed plan consider the debtor’s real financial capacity? \\
    \vspace{0.5em}
    2. Emotion Support Score (0–10) \\
    Evaluate the collector’s emotional intelligence and supportiveness based on: \\
    - Identification: Ability to uncover the debtor’s underlying struggles. \\
    - Comforting: Use of empathetic, warm, and compassionate responses. \\
    - Experience: Use of relevant analogies or shared experiences to build connection. \\
    - Emotional Impact: Final emotional state of the debtor (low to extreme negative emotions). \\
    \vspace{0.5em}
    3. Communication Ability Score (0–10) \\
    Assess the collector’s conversational effectiveness based on: \\
    - Consistency: Logical flow and contextual coherence of responses. \\
    - Role-adherence: Staying professionally consistent without contradictions. \\
    - Expression: Linguistic variety, clarity, and engagement. \\
    - Humanness: Naturalness and human-like quality of responses. \\
    \vspace{0.5em}
    ******** \\
    You should output it in JSON format: \\
    \{"satisfaction\_score": xxx, "satisfaction\_reason": "Brief explanation.", "emotion\_support\_score": xxx, "emotion\_support\_reason": "Brief explanation.", "communication\_ability\_score": xxx, "communication\_ability\_reason": "Brief explanation."\} \\
    Please note: You must output only one JSON object in this exact format. No additional text. Answer in English. \\
    \hline
    \end{tabular}
    \caption{The complete prompt of \textbf{Interaction Experience Evaluation}.}
\end{table*}

\end{document}